# Algorithms for Speech Recognition and Language Processing

*Mehryar Mohri*

AT&T Laboratories

mohri@research.att.com

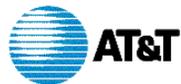

*Michael Riley*

AT&T Laboratories

riley@research.att.com

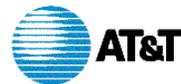

*Richard Sproat*

Bell Laboratories

rws@bell-labs.com

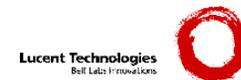

Joint work with

*Emerald Chung, Donald Hindle, Andrej Ljolje, Fernando Pereira*

*Tutorial presented at COLING'96, August 3rd, 1996.*

# Introduction (1)

Text and speech processing: hard problems

- Theory of automata

- Appropriate level of abstraction

- Well-defined algorithmic problems

# Introduction (2)

Three Sections:

- Algorithms for text and speech processing (2h)

- Speech recognition (2h)

- Finite-state methods for language processing (2h)

# PART I
# Algorithms for Text and Speech Processing

*Mehryar Mohri*

AT&T Laboratories

mohri@research.att.com

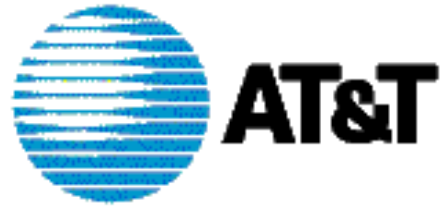

August 3rd, 1996

# Definitions: finite automata (1)

$A = (\Sigma, Q, \delta, I, F)$

- Alphabet $\Sigma$,
- Finite set of states $Q$,
- Transition function $\delta$: $Q \times \Sigma \to 2^Q$,
- $I \subseteq Q$ set of initial states,
- $F \subseteq Q$ set of final states.

$A$ recognizes $L(A) = \{w \in \Sigma^* : \delta(I, w) \cap F \neq \emptyset\}$
(Hopcroft and Ullman, 1979; Perrin, 1990)

**Theorem 1** *(Kleene, 1965). A set is regular (or rational) iff it can be recognized by a finite automaton.*

# Definitions: finite automata (2)

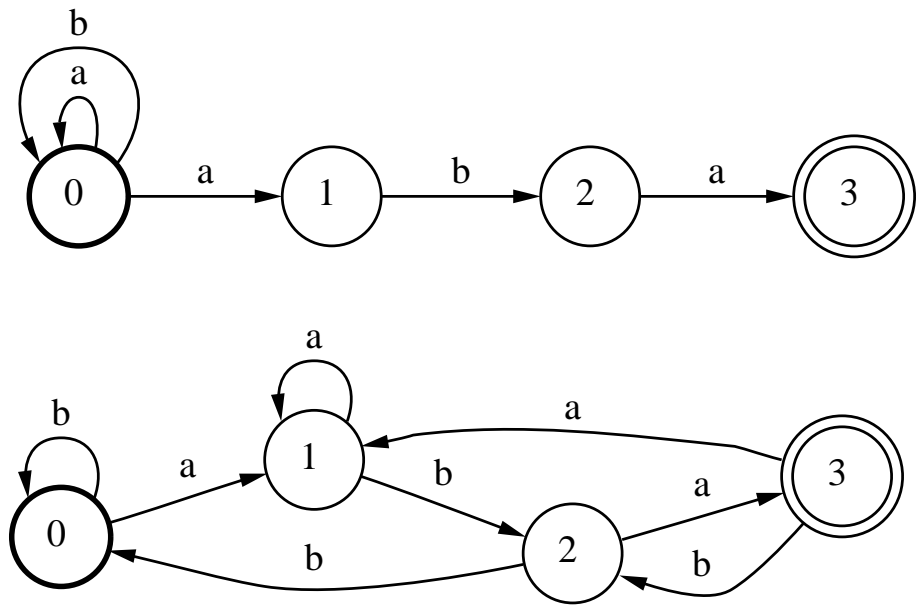

Figure 1: $L(A) = \Sigma^* aba$.

# Definitions: weighted automata (1)

$A = (\Sigma, Q, \lambda, \delta, \sigma, \rho, I, F)$

- $(\Sigma, Q, \delta, I, F)$ is an automaton,

- Initial output function $\lambda$,

- Output function $\sigma$: $Q \times \Sigma \times Q \to K$,

- Final output function $\rho$,

- Function $f : \Sigma^* \to (K, +, \cdot)$ associated with $A$:
  $$\forall u \in Dom(f), f(u) = \sum_{(i,q) \in I \times (\delta(i,u) \cap F)} (\lambda(i) \cdot \sigma(i, u, q) \cdot \rho(q)).$$

# Definitions: weighted automata (2)

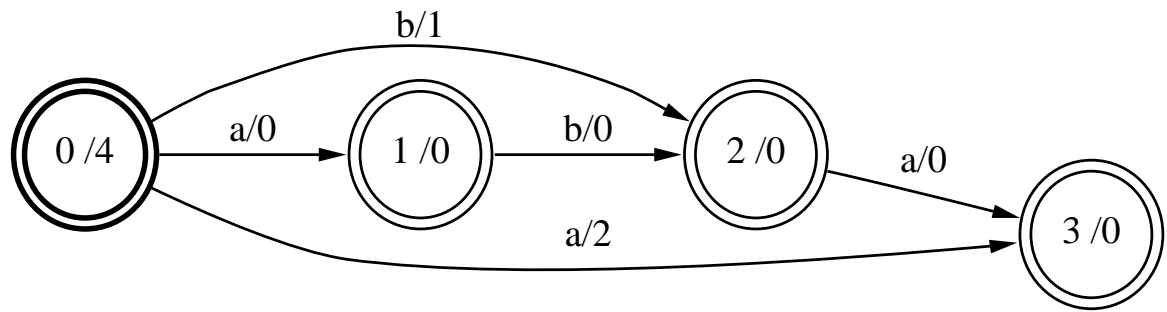

Figure 2: Index of $t = aba$.

# Definitions: rational power series

- *Power series*: functions mapping $\Sigma^*$ to a semiring $(K, +, \cdot)$
  - Notation: $S = \sum_{w \in \Sigma^*} (S, w) w$, $(S, w)$: coefficients
  - Support: $supp(S) = \{w \in \Sigma^* : (S, w) \neq 0\}$
  - Sum: $(S + T, w) = (S, w) + (T, w)$
  - Star: $S^* = \sum_{n \geq 0} S^n$
  - Product: $(ST, w) = \sum_{uv = w \in \Sigma^*} (S, u)(T, v)$

- *Rational power series*: closure under rational operations of polynomials (polynomial power series) (Salomaa and Soittola, 1978; Berstel and Reutenauer, 1988)

**Theorem 2** *(Schützenberger, 1961). A power series is rational iff it can be represented by a weighted finite automaton.*

# Definitions: transducers (1)

$T = (\Sigma, \Delta, Q, \delta, \sigma, I, F)$

- Finite alphabets $\Sigma$ and $\Delta$,
- Finite set of states $Q$,
- Transition function $\delta$: $Q \times \Sigma \to 2^Q$,
- Output function $\sigma$: $Q \times \Sigma \times Q \to \Sigma^*$,
- $I \subseteq Q$ set of initial states,
- $F \subseteq Q$ set of final states.

$T$ defines a relation:
$$R(T) = \{(u,v) \in (\Sigma^*)^2 : v \in \bigcup_{q \in (\delta(I,u) \cap F)} \sigma(I, u, q)\}$$

# Definitions: transducers (2)

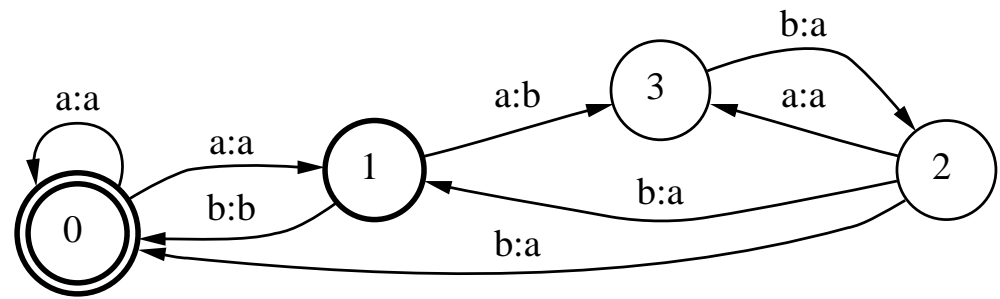

Figure 3: Fibonacci normalizer ($[abb \rightarrow baa] \circ [baa \leftarrow abb]$).

## Definitions: weighted transducers

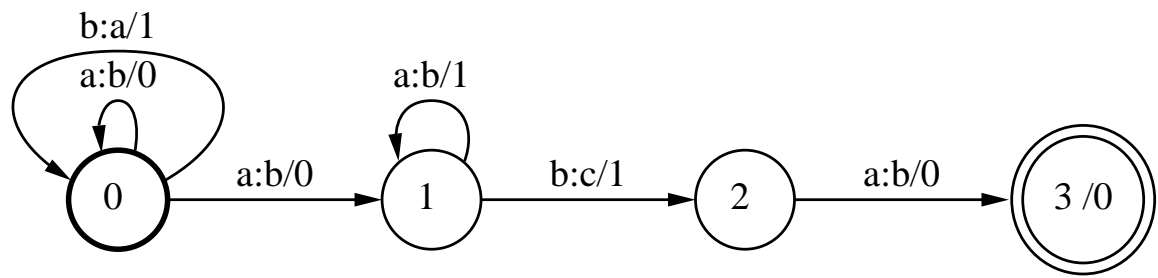

Figure 4: Example, $aaba \to (bbcb, (0 \odot 0 \odot 1 \odot 0) \oplus (0 \odot 1 \odot 1 \odot 0))$.

$(\min, +):\quad aaba \to \min\{1, 2\} = 1$

$(+, \cdot):\quad aaba \to 0 + 0 = 0$

# Composition: Motivation (1)

- Construction of complex sets or functions from more elementary ones

- Modular (modules, distinct linguistic descriptions)

- *On-the-fly* expansion

# Composition: Motivation (2)

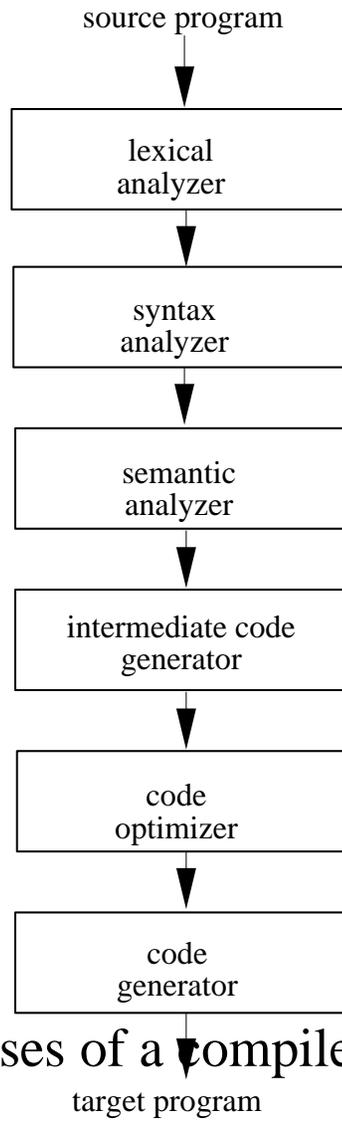

Figure 5: Phases of a compiler (Aho *et al.*, 1986).

## Composition: Motivation (3)

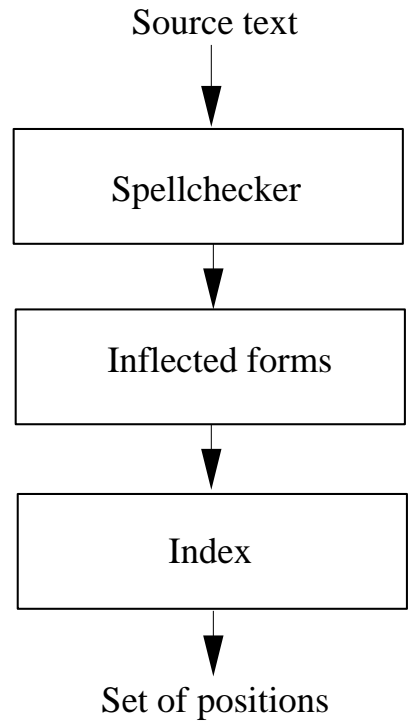

Figure 6: Complex indexation.

# Composition: Example (1)

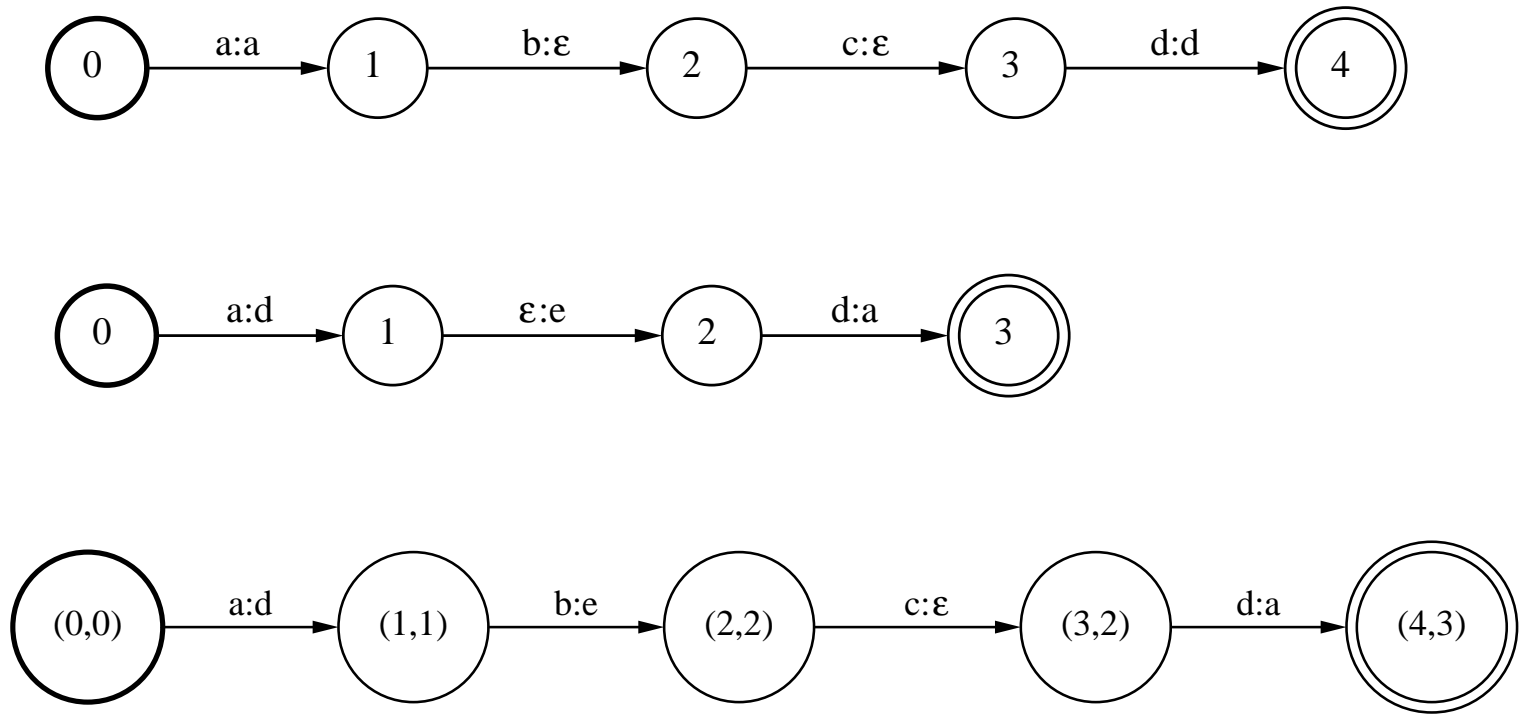

Figure 7: Composition of transducers.

# Composition: Example (2)

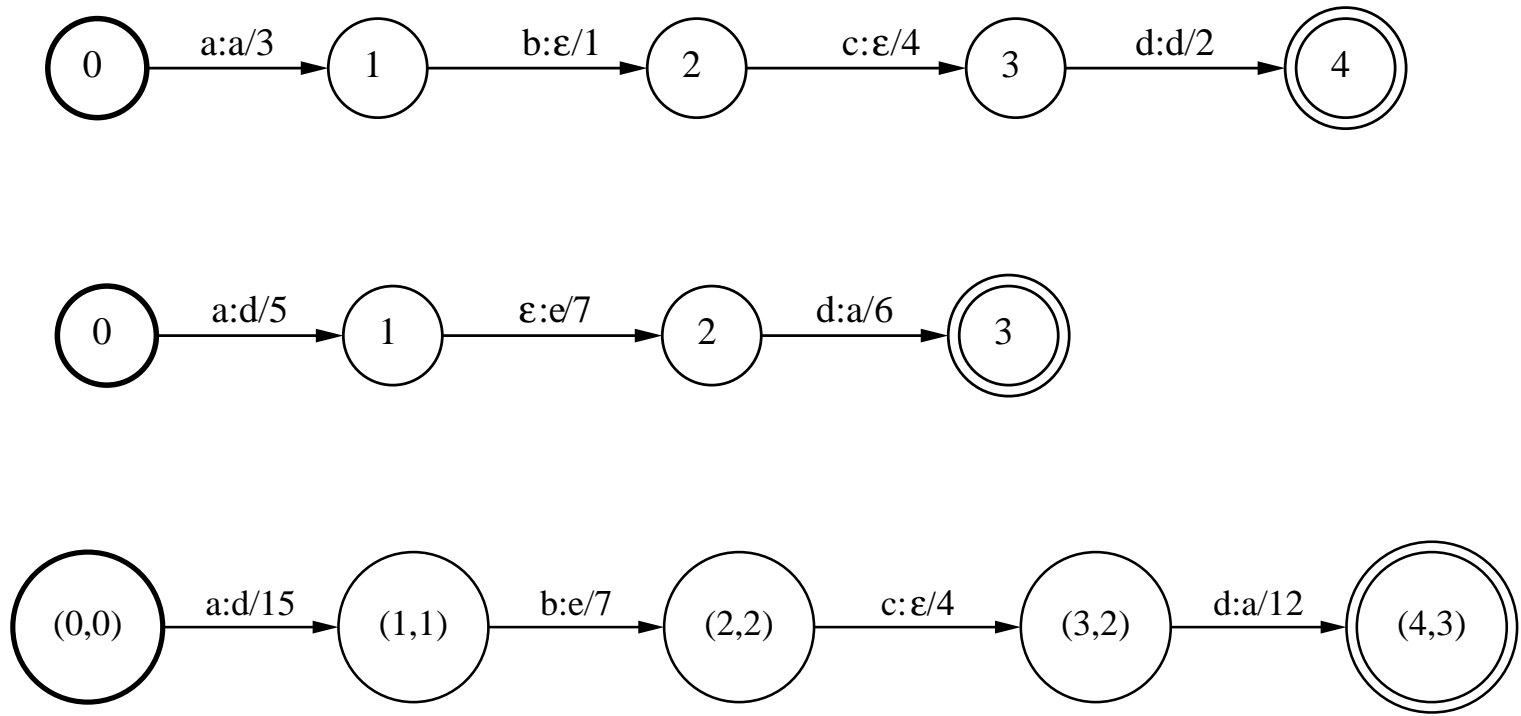

Figure 8: Composition of weighted transducers $(+, \cdot)$.

# Composition: Algorithm (1)

- Construction of pairs of states
  - Match: $q_1 \xrightarrow{a:b/w_1} q_1'$ and $q_2 \xrightarrow{b:c/w_2} q_2'$
  - Result: $(q_1, q_2) \xrightarrow{a:c/(w_1 \odot w_2)} (q_1', q_2')$

- Elimination of $\epsilon$-paths redundancy: filter

- Complexity: quadratic

- *On-the-fly* implementation

# Composition: Algorithm (2)

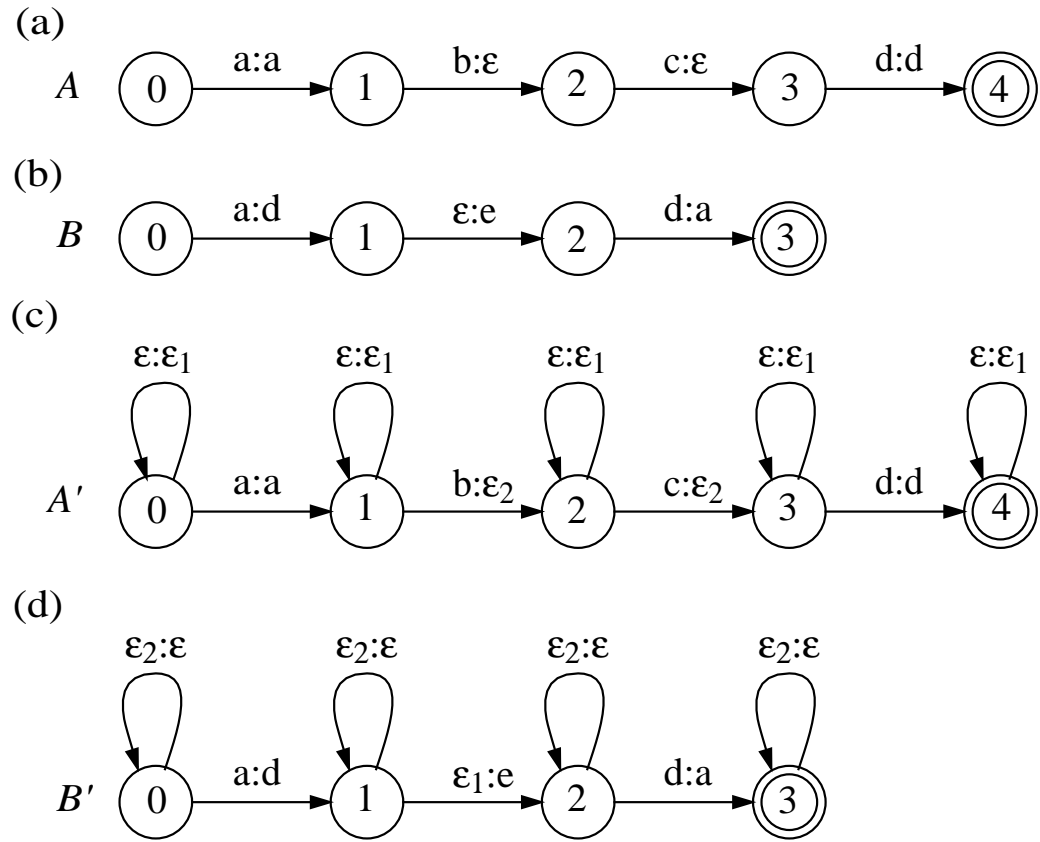

Figure 9: Composition of weighted transducers with $\epsilon$-transitions.

## Composition: Algorithm (3)

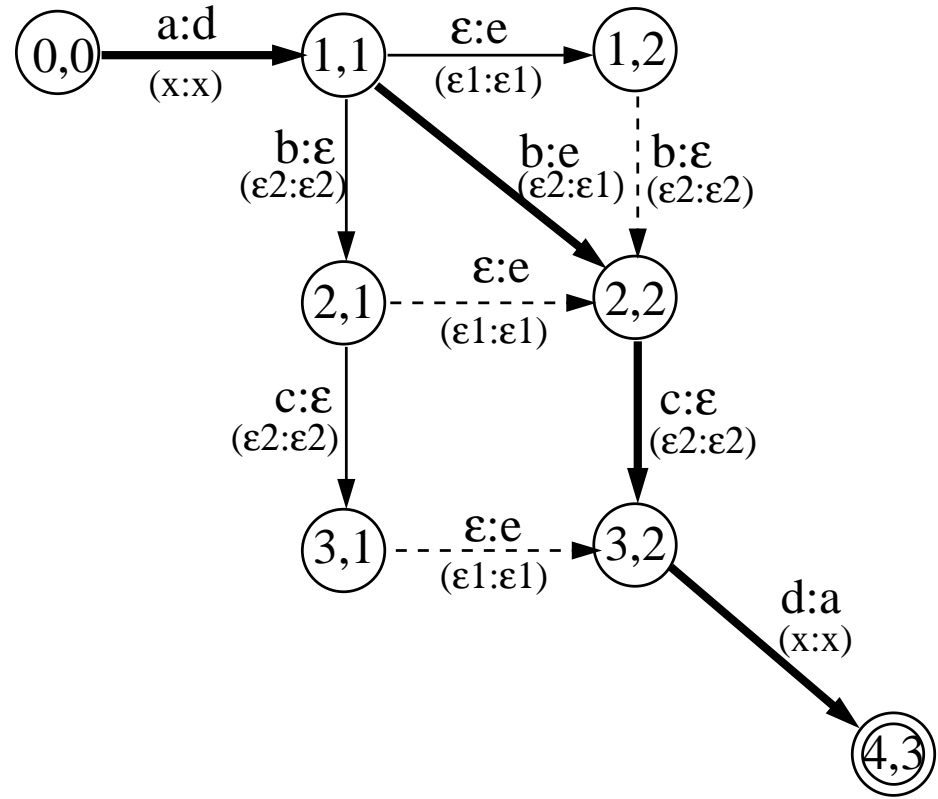

Figure 10: Redundancy of $\epsilon$-paths.

# Composition: Algorithm (4)

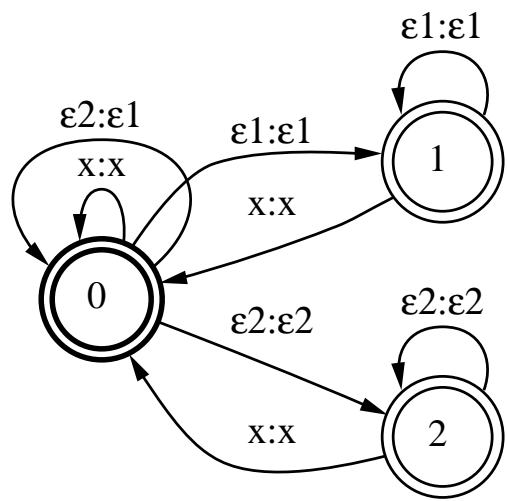

Figure 11: Filter for efficient composition.

# Composition: Theory

- Transductions (Elgot and Mezei, 1965; Eilenberg, 1974 1976; Berstel, 1979).

- **Theorem 3** *Let $\tau_1$ and $\tau_2$ be two (weighted) (automata + transducers), then $(\tau_1 \circ \tau_2)$ is a (weighted) (automaton + transducer).*

- Efficient composition of weighted transducers (Mohri, Pereira, and Riley, 1996).

- Works with any semiring

- Intersection: composition of automata (weighted).

# Intersection: Example

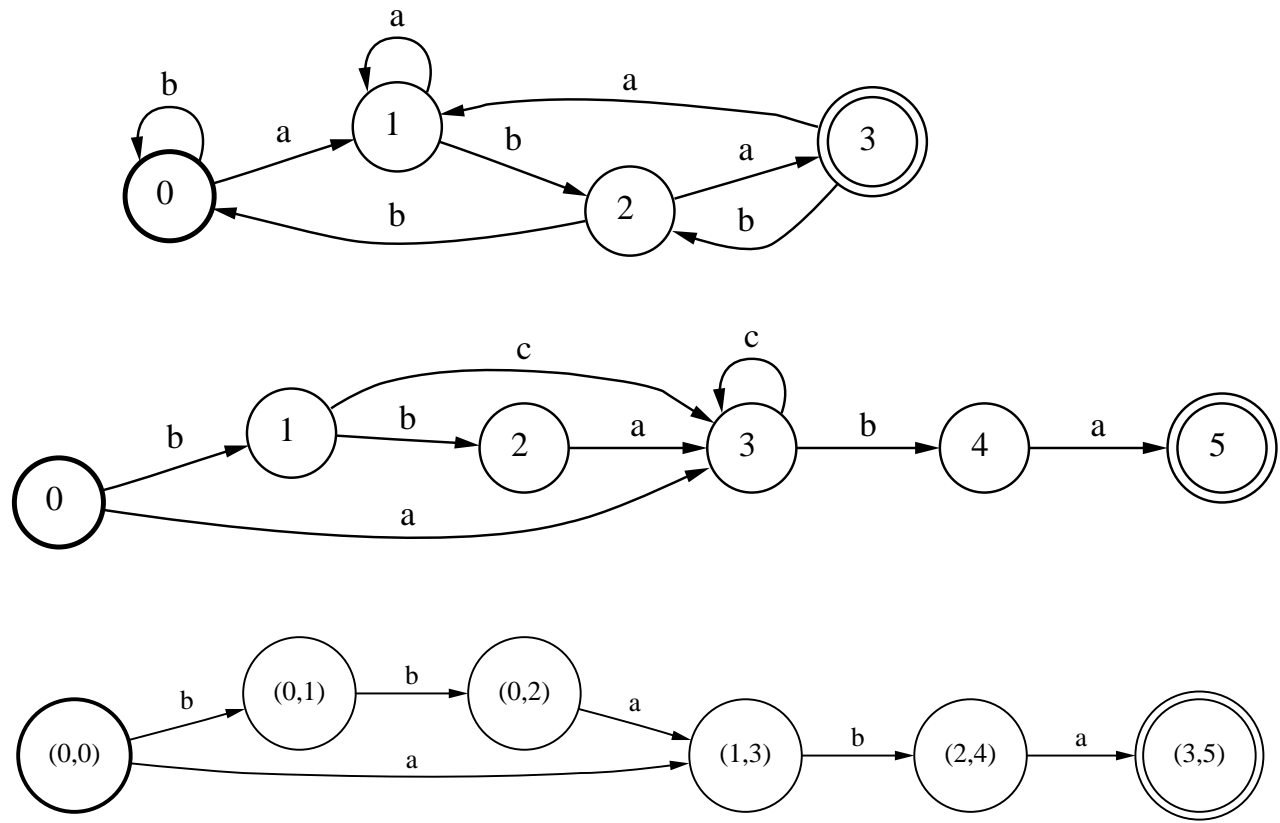

Figure 12: Intersection of automata.

# Union: Example

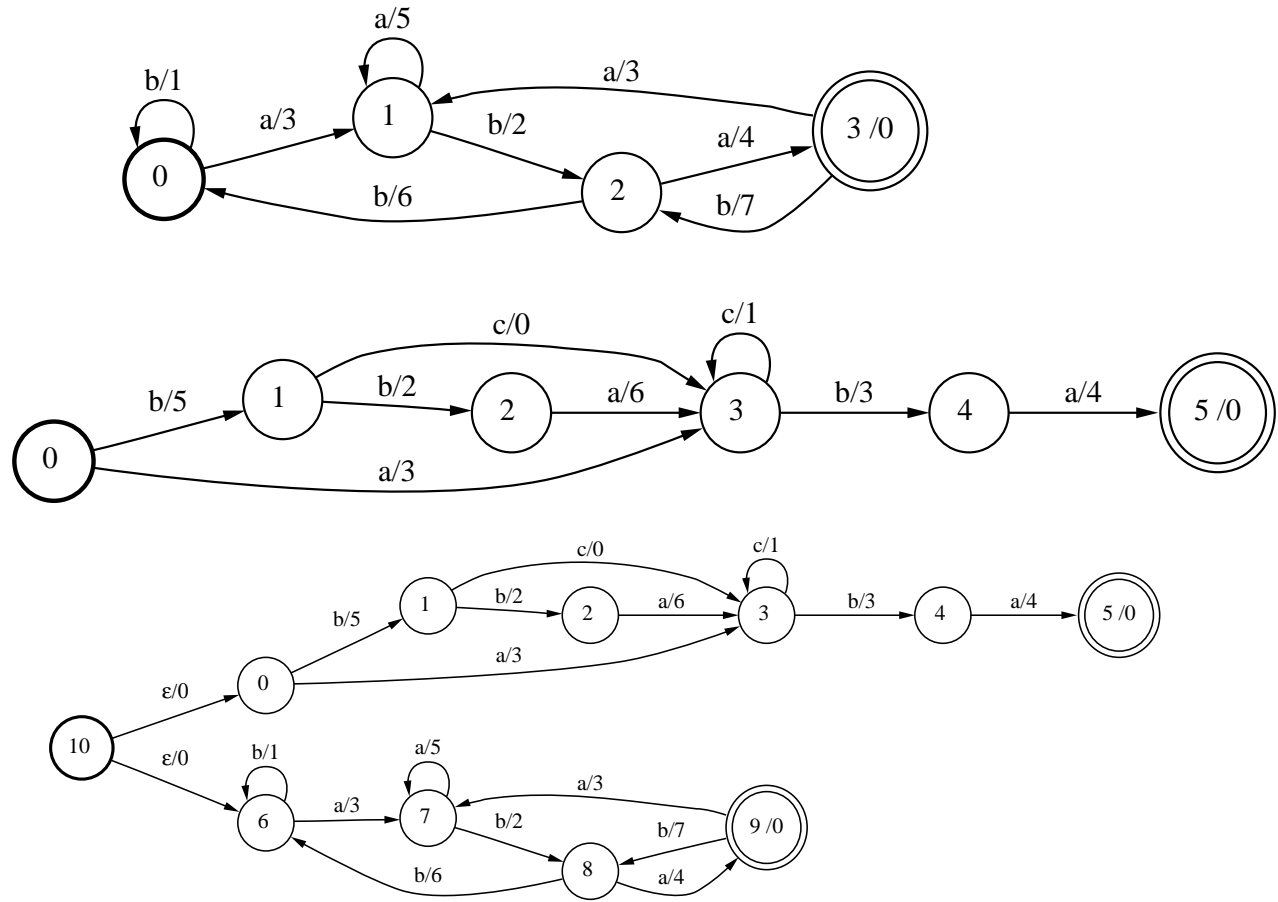

Figure 13: Union of weighted automata (min, +).

# Determinization: Motivation (1)

- Efficiency of use (time)

- Elimination of redundancy

- No loss of information ($\neq$ pruning)

# Determinization: Motivation (2)

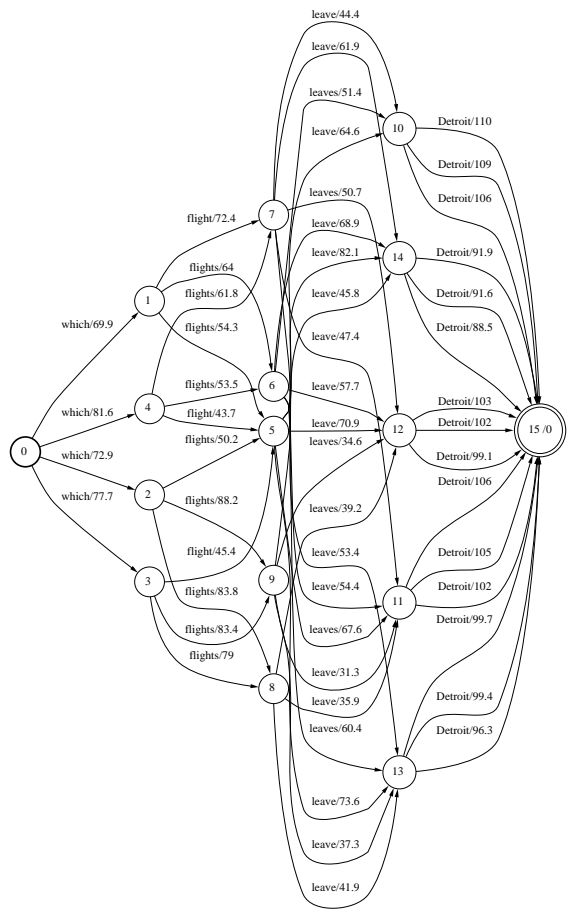

Figure 14: *Toy* language model (16 states, 53 transitions, 162 paths).

# Determinization: Motivation (3)

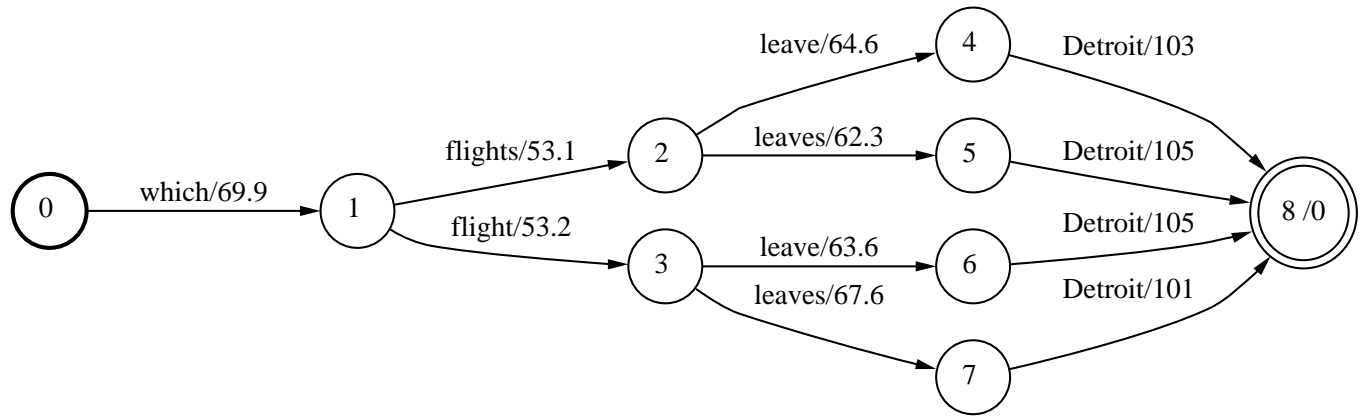

Figure 15: Determinized *language model* (9 states, 11 transitions, 4 paths).

# Determinization: Example (1)

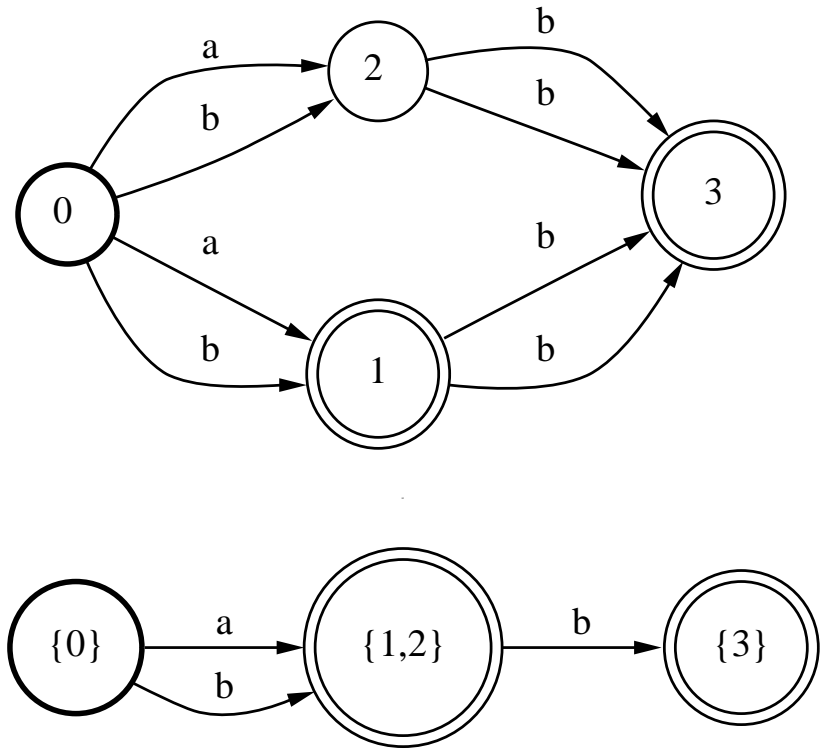

Figure 16: Determinization of automata.

## Determinization: Example (2)

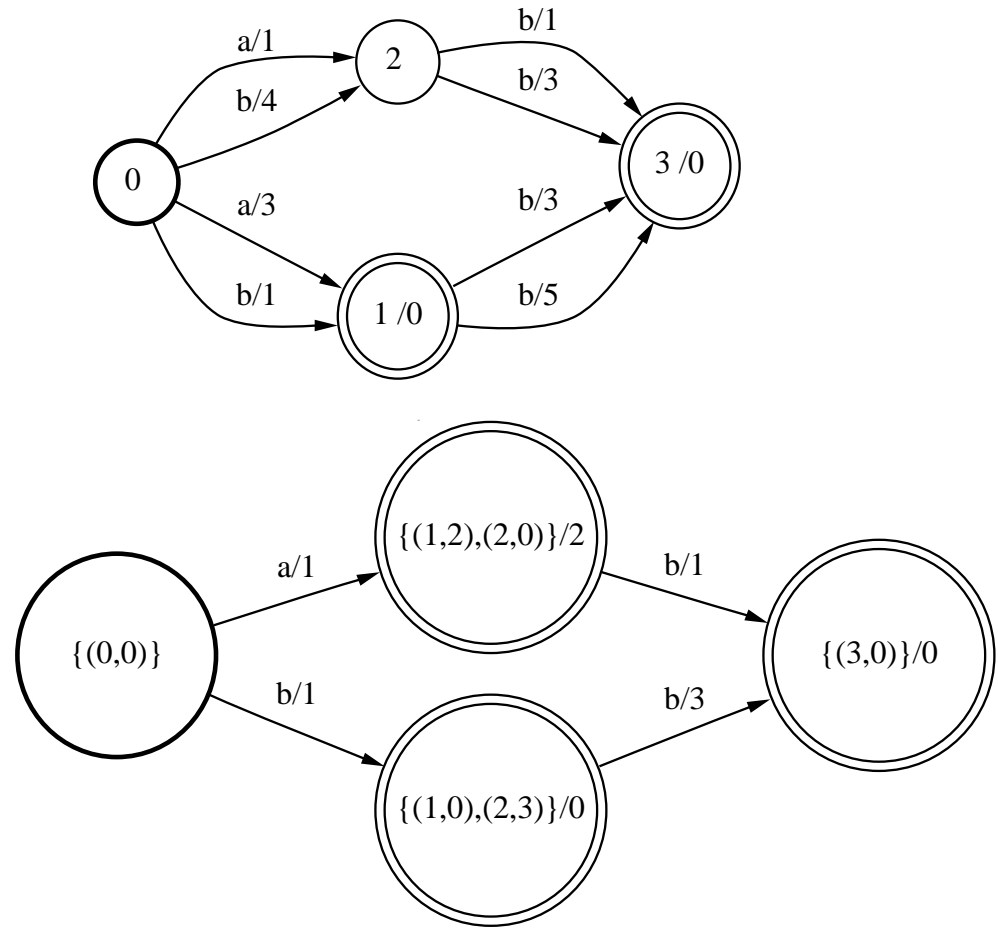

Figure 17: Determinization of weighted automata (min, +).

# Determinization: Example (3)

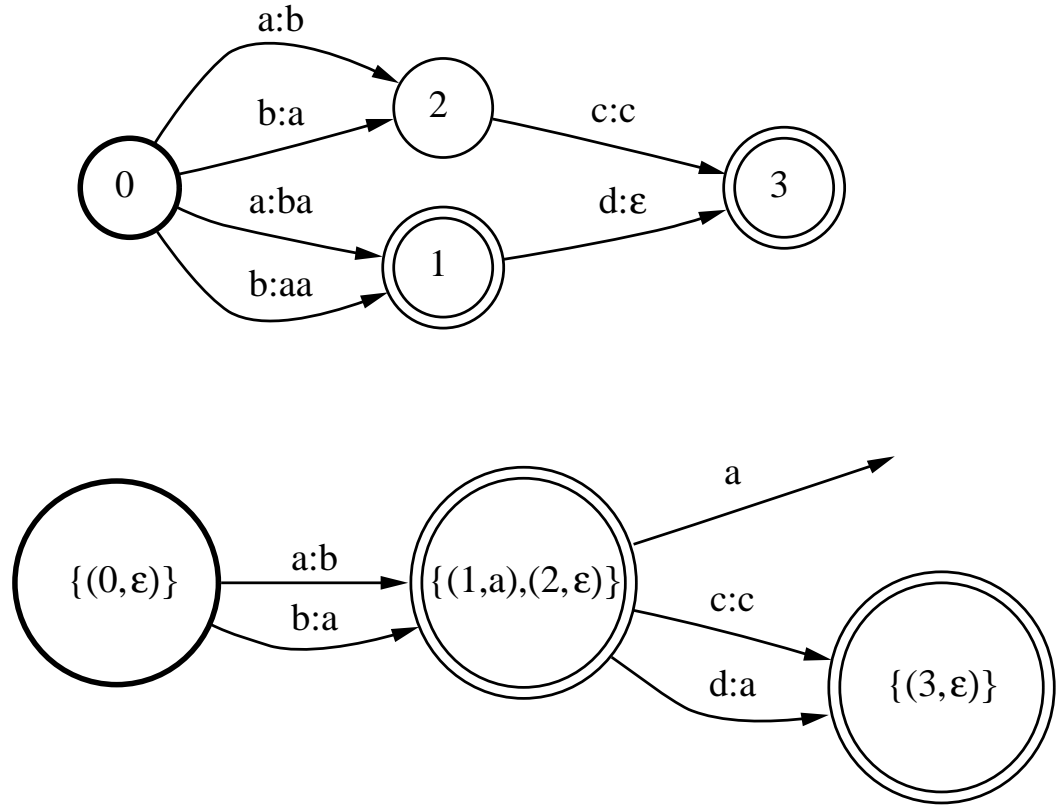

Figure 18: Determinization of transducers.

# Determinization: Example (4)

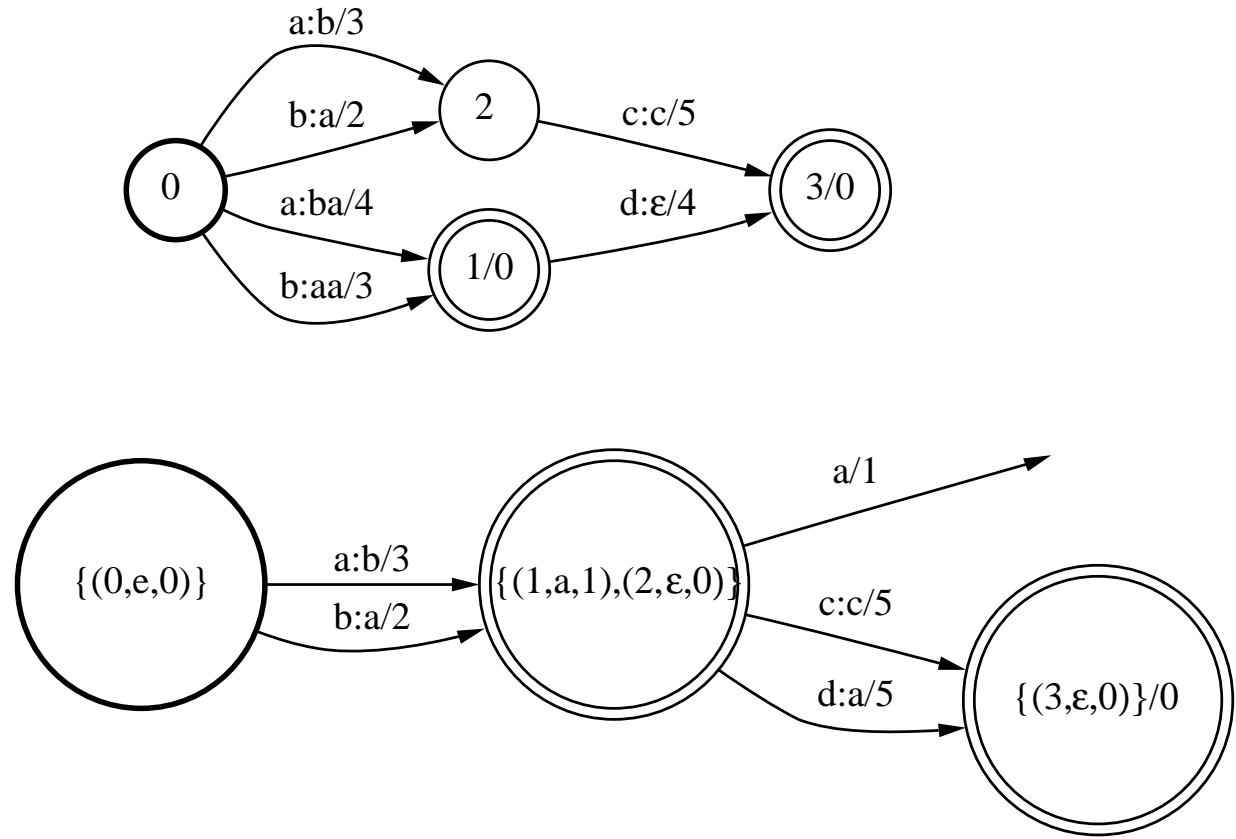

Figure 19: Determinization of weighted transducers (min, +).

# Determinization: Algorithm (1)

- Generalization of the classical algorithm for automata
  - Powerset construction
  - Subsets made of (state, weight) or (state, string, weight)

- Applies to subsequentiable weighted automata and transducers

- Time and space complexity: exponential (polynomial w.r.t. size of the result)

- *On-the-fly* implementation

# Determinization: Algorithm (2)

## Conditions of applications

- *Twin* states: $q$ and $q'$ are twin states iff:
  - If: they can be reached from the initial states by the same input string $u$
  - Then: cycles at $q$ and $q'$ with the same input string $v$ have the same output value

- **Theorem 4** *(Choffrut, 1978; Mohri, 1996a) Let $\tau$ be an unambiguous weighted automaton (transducer, weighted transducer), then $\tau$ can be determinized iff it has the twin property.*

- **Theorem 5** *(Mohri, 1996a) The twin property can be tested in polynomial time.*

# Determinization: Theory

- Determinization of automata

    - General case (Aho, Sethi, and Ullman, 1986)

    - Specific case of $\Sigma^* \alpha$: failure functions (Mohri, 1995)

- Determinization of transducers, weighted automata, and weighted transducers

    - General description, theory and analysis (Mohri, 1996a; Mohri, 1996b)

    - Conditions of application and test algorithm

    - Acyclic weighted transducers or transducers admit determinization

- Can be used with other semirings (ex: $(\mathcal{R}, +, \cdot)$)

# Local determinization: Motivation

- Time efficiency

- Reduction of redundancy

- Control of the resulting size (flexibility)

- Equivalent function (or equal set)

- No loss of information

# Local determinization: Example

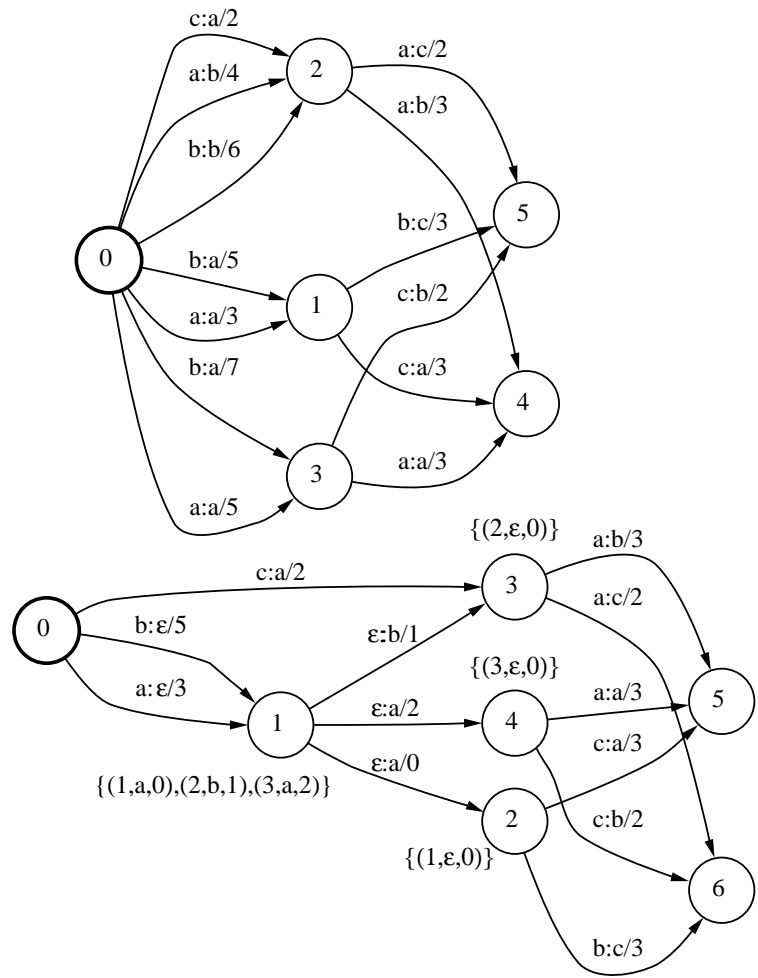

Figure 20: Local determinization of weighted transducers $(\min, +)$.

# Local determinization: Algorithm

- Predicate, ex: $(P)\,(out-degree(q) > k)$

- $k$: threshold parameter

- Local: $Dom(det) = \{q : P(q)\}$

- *Determinization* only for $q \in Dom(det)$

- *On-the-fly* implementation

- Complexity $O(|Dom(det)| \cdot \max_{q \in Q}(out-degree(q)))$

# Local determinization: theory

- Various choices of predicate (constraint: local)

- Definition of parameters

- Applies to all automata, weighted automata, transducers, and weighted transducers

- Can be used with other semirings (ex: $(\mathcal{R}, +, \cdot)$)

# Minimization: Motivation

- Space efficiency

- Equivalent function (or equal set)

- No loss of information ($\neq$ pruning)

# Minimization: Motivation (2)

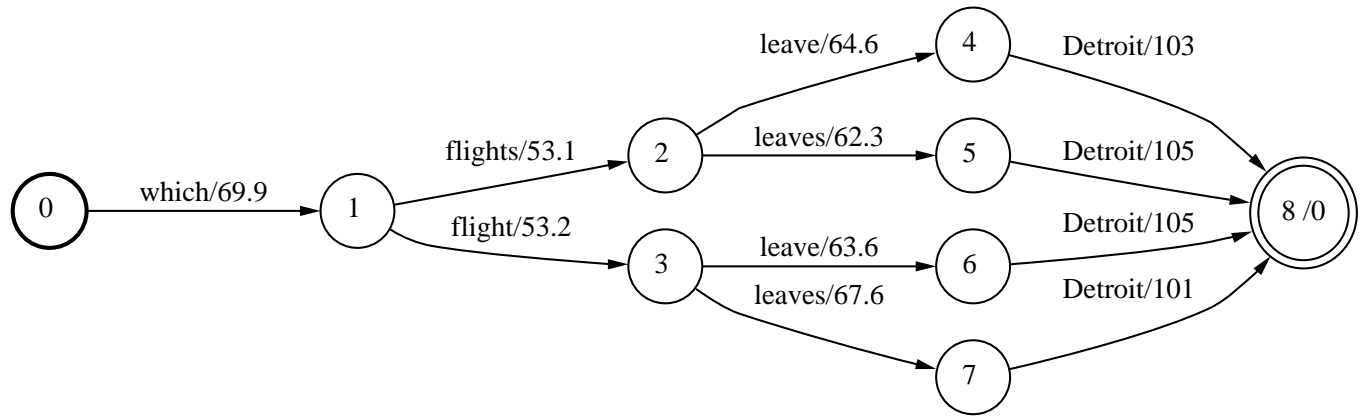

Figure 21: Determinized *language model*.

# Minimization: Motivation (3)

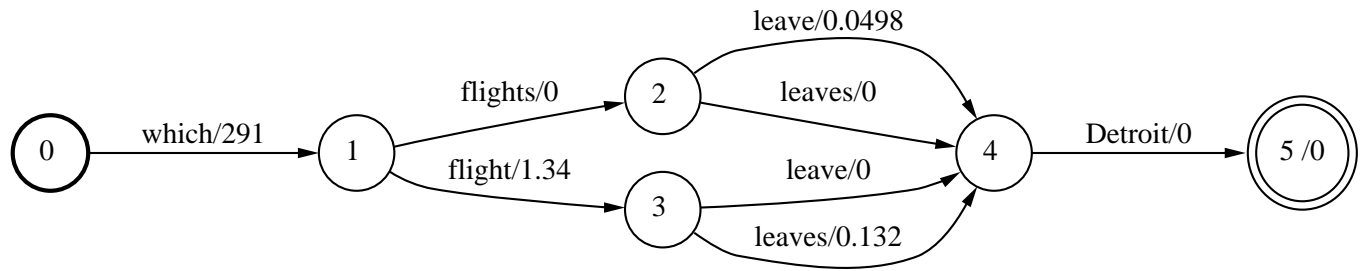

Figure 22: Minimized *language model*.

# Minimization: Example (1)

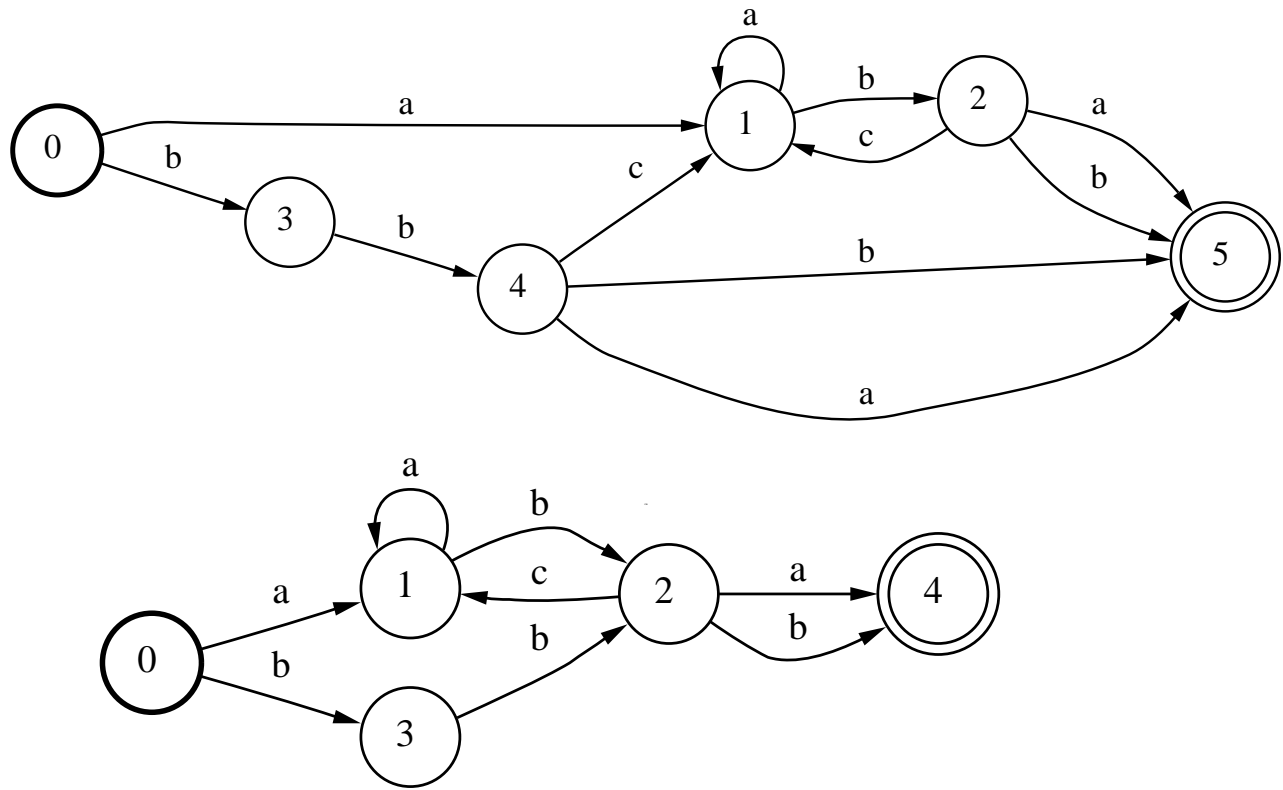

Figure 23: Minimization of automata.

# Minimization: Example (2)

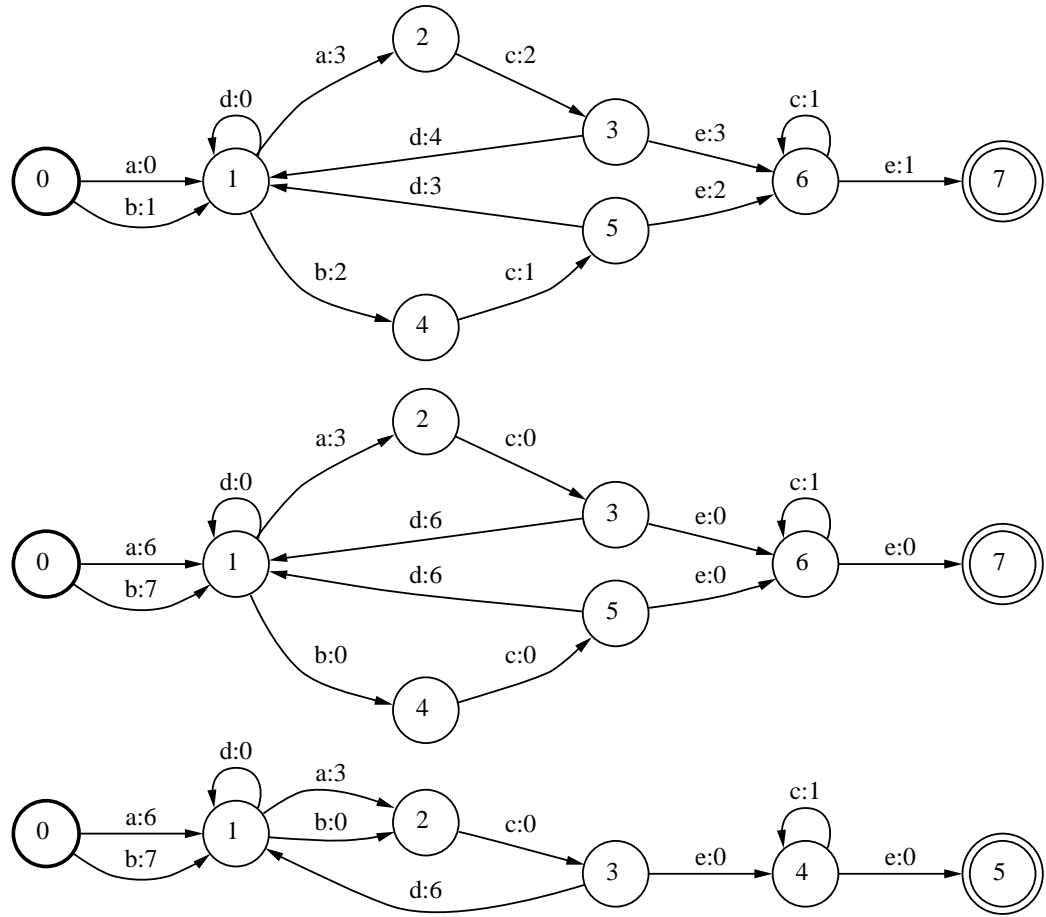

Figure 24: Minimization of weighted automata (min, +).

# Minimization: Example (3)

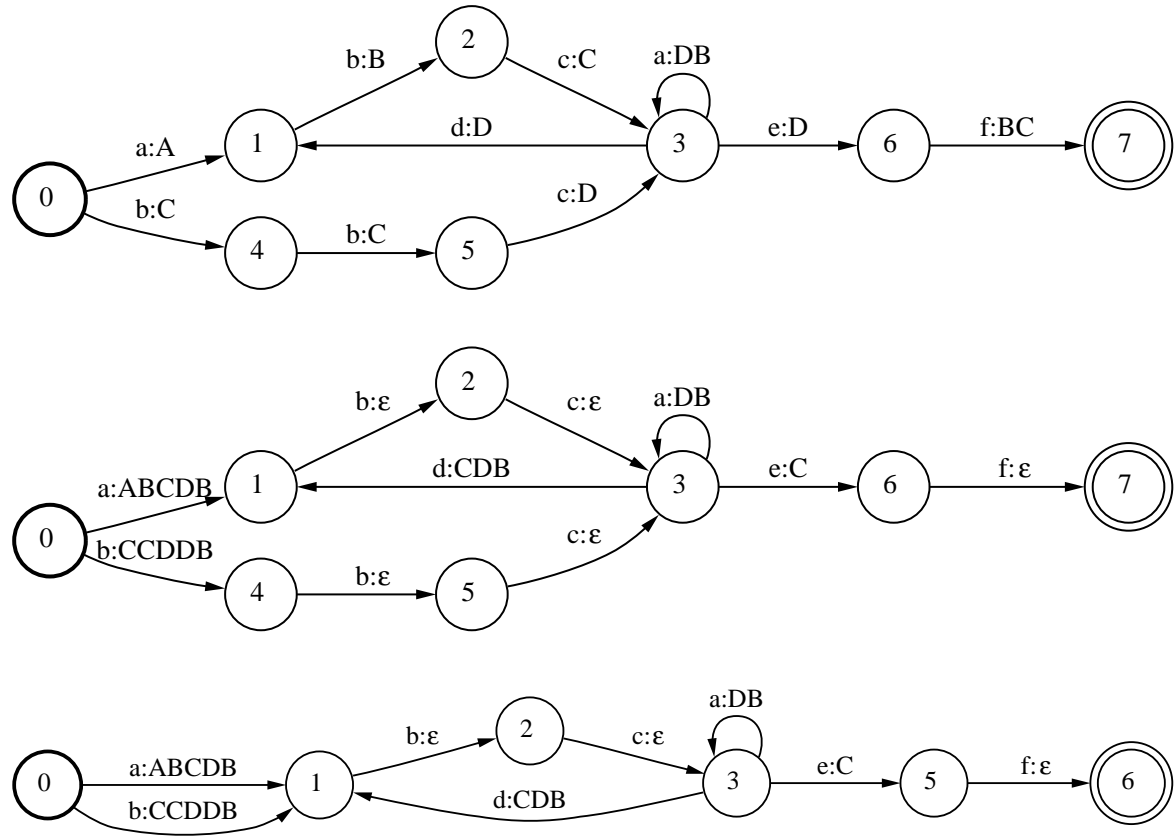

Figure 25: Minimization of transducers.

# Minimization: Example (4)

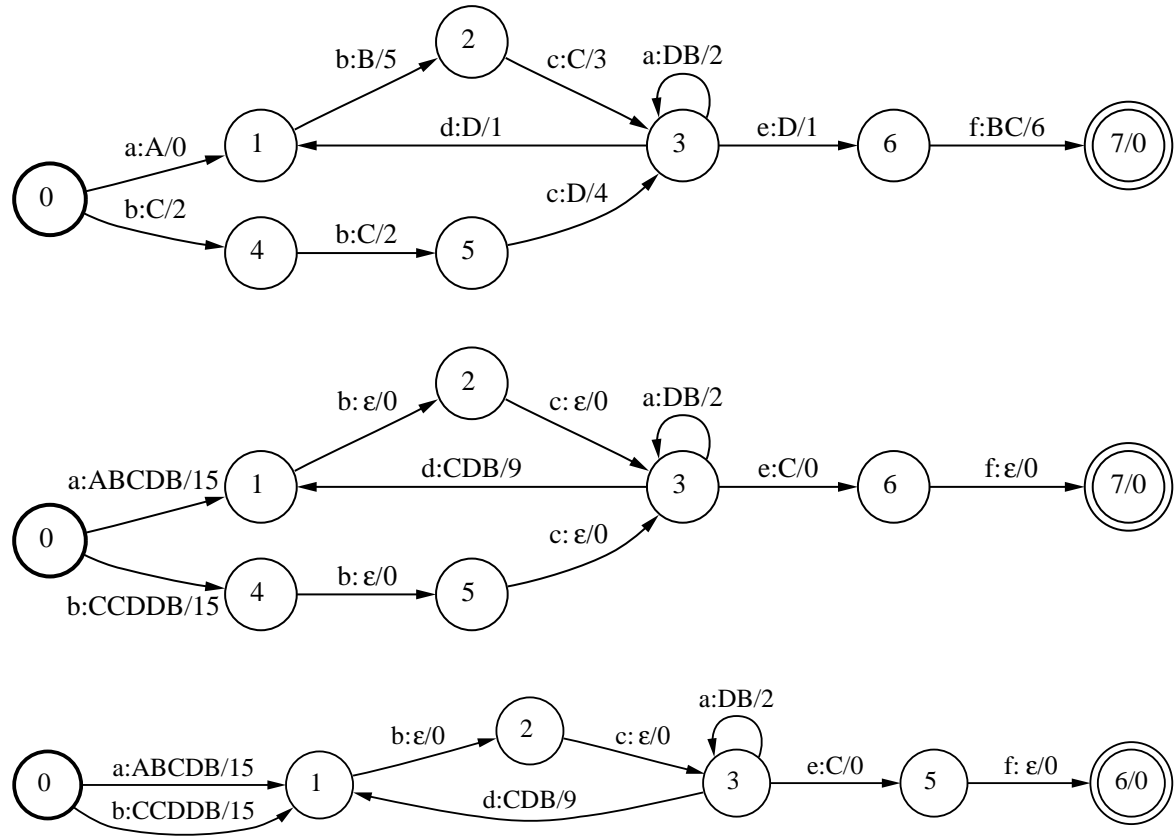

Figure 26: Minimization of weighted transducers (min, +).

# Minimization: Algorithm (1)

- Two steps

  - *Pushing* or *extraction* of strings or weights towards initial state

  - Classical minimization of automata, (input,ouput) considered as a single label

- Algorithm for the first step

  - Transducers: specific algorithm

  - Weighted automata: shortest-paths algorithms

# Minimization: Algorithm (2)

- Complexity
  - E: set of transitions
  - S: sum of the lengths of output strings
  - the longest of the longest common prefixes of the output paths leaving each state

| Type | General | Acyclic |
| --- | --- | --- |
| Automata | $O(\|E\| \cdot \log(\|Q\|))$ | $O(\|Q\| + \|E\|)$ |
| Weighted automata | $O(\|E\| \cdot \log(\|Q\|))$ | $O(\|Q\| + \|E\|)$ |
| Transducers | $O(\|Q\| + \|E\| \cdot (\log \|Q\| + \|P_{max}\|))$ | $O(S + \|E\| + \|Q\| + (\|E\| - (\|Q\| - \|F\|)) \cdot \|P_{max}\|)$ |

# Minimization: Theory

- Minimization of automata (Aho, Hopcroft, and Ullman, 1974; Revuz, 1991)

- Minimization of transducers (Mohri, 1994)

- Minimization of weighted automata (Mohri, 1996a)
  - Minimal number of transitions
  - Test of equivalence

- Standardization of power series (Schützenberger, 1961)
  - Works only with fields
  - Creates too many transitions

# Conclusion (1)

- Theory
  - Rational power series
  - Weighted automata and transducers

- Algorithms
  - General (various semirings)
  - Efficiency (used in practice, large sizes)

# Conclusion (2)

- Applications
    - Text processing
      (spelling checkers, pattern-matching, indexation, OCR)
    - Language processing
      (morphology, phonology, syntax, language modeling)
    - Speech processing (speech recognition, text-to-speech synthesis)
    - Computational biology (matching with errors)
    - Many other applications

# PART II
# Speech Recognition

*Michael Riley*

AT&T Laboratories

riley@research.att.com

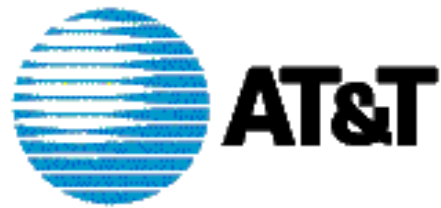

August 3rd, 1996

# Overview

- The speech recognition problem

- Acoustic, lexical and grammatical models

- Finite-state automata in speech recognition

- Search in finite-state automata

# Speech Recognition

*Given an utterance, find its most likely written transcription.*

Fundamental ideas:

- Utterances are built from sequences of units

- Acoustic correlates of a unit are affected by surrounding units

- Units combine into units at a higher level — phones → syllables → words

- Relationships between levels can be modeled by weighted graphs — we use *weighted finite-state transducers*

- *Recognition: find the best path in a suitable product graph*

# Levels of Speech Representation

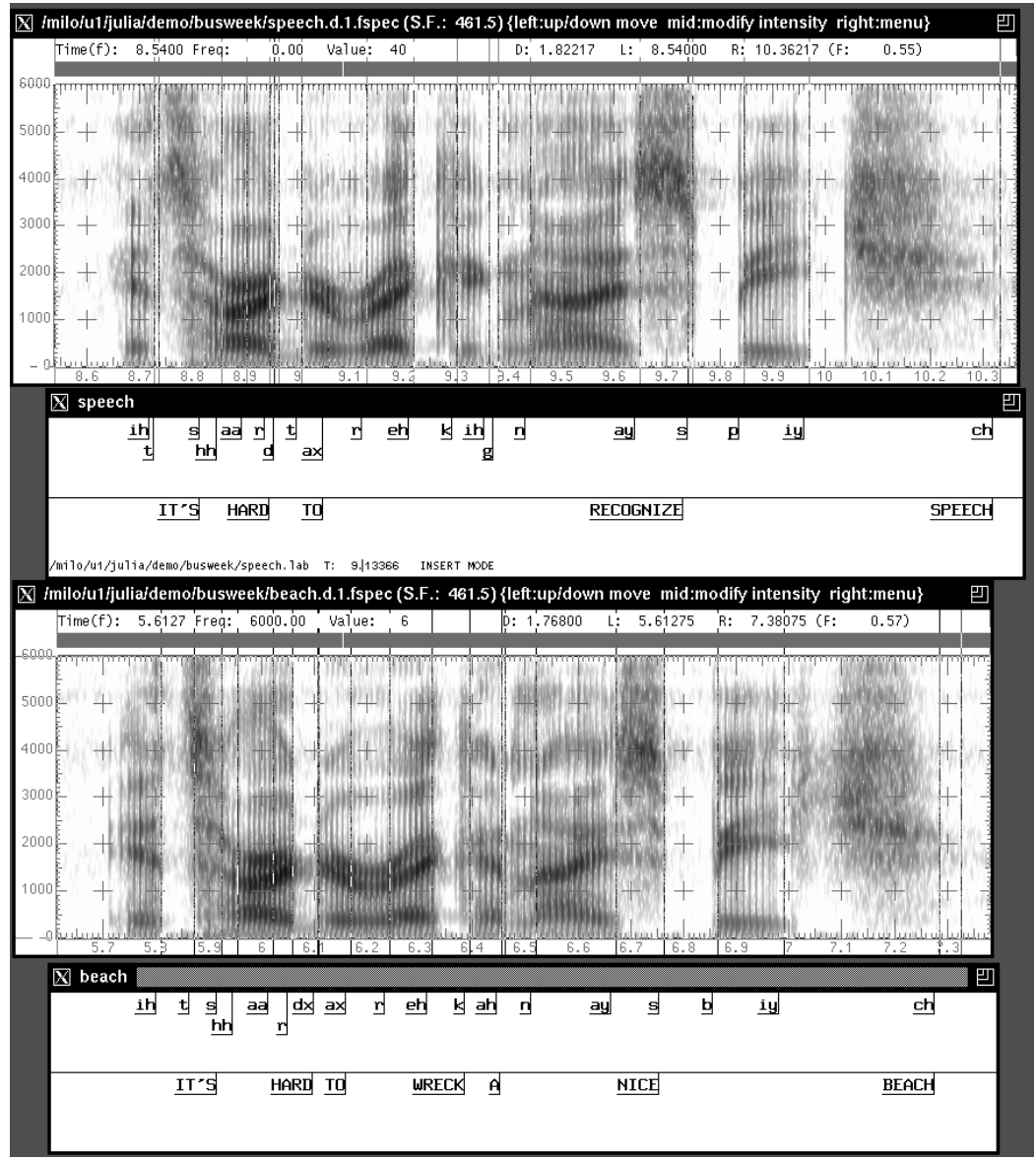

# Maximum A Posteriori Decoding

Overall analysis [4, 57]:

- *Acoustic observations:* parameter vectors derived by local spectral analysis of the speech waveform at regular (e.g. 10msec) intervals

- Observation sequence $\mathbf{o}$

- Transcriptions $\mathbf{w}$

- Probability $P(\mathbf{o}|\mathbf{w})$ of observing $\mathbf{o}$ when $\mathbf{w}$ is uttered

- *Maximum a posteriori decoding*:

$$\hat{\mathbf{w}} = \operatorname*{argmax}_{\mathbf{w}} P(\mathbf{w}|\mathbf{o}) = \operatorname*{argmax}_{\mathbf{w}} \frac{P(\mathbf{o}|\mathbf{w})P(\mathbf{w})}{P(\mathbf{o})}$$

$$= \operatorname*{argmax}_{\mathbf{w}} \underbrace{P(\mathbf{o}|\mathbf{w})}_{\text{generative model}} \underbrace{P(\mathbf{w})}_{\text{language model}}$$

# Generative Models of Speech

Typical decomposition of $P(\mathbf{o}|\mathbf{w})$ into conditionally-independent mappings between levels:

- Acoustic model $P(\mathbf{o}|\mathbf{p})$ : phone sequences $\rightarrow$ observation sequences. Detailed model:

    - $P(o|d)$ : distributions $\rightarrow$ observation vectors — *symbolic $\rightarrow$ quantitative*

    - $P(\mathbf{d}|m)$ : context-dependent phone models $\rightarrow$ distribution sequences

    - $P(\mathbf{m}|\mathbf{p})$ : phone sequences $\rightarrow$ model sequences

- Pronunciation model $P(\mathbf{p}|\mathbf{w})$ : word sequences $\rightarrow$ phone sequences

- Language model $P(\mathbf{w})$ : word sequences

# Recognition Cascades: General Form

- Multistage cascade:

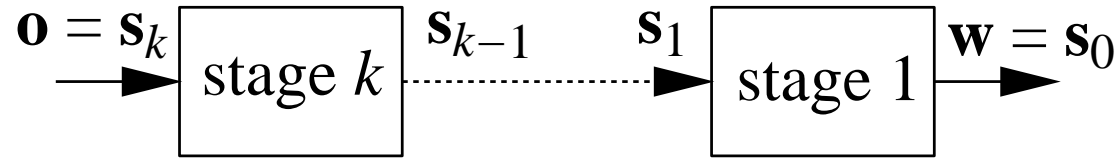

Find $\mathbf{s}_0$ maximizing

$$P(\mathbf{s}_0, \mathbf{s}_k) = P(\mathbf{s}_k|\mathbf{s}_0)P(\mathbf{s}_0) = P(\mathbf{s}_0) \sum_{\mathbf{s}_1,\ldots,\mathbf{s}_{k-1}} \prod_{1 \leq j \leq k} P(\mathbf{s}_j|\mathbf{s}_{j-1})$$

- "Viterbi" approximation:

$$\begin{aligned} \text{Cost}(\mathbf{s}_0, \mathbf{s}_k) &= \text{Cost}(\mathbf{s}_k|\mathbf{s}_0) + \text{Cost}(\mathbf{s}_0) \\ \text{Cost}(\mathbf{s}_k|\mathbf{s}_0) &\approx \min_{\mathbf{s}_1,\ldots,\mathbf{s}_{k-1}} \sum_{1 \leq j \leq k} \text{Cost}(\mathbf{s}_j|\mathbf{s}_{j-1}) \end{aligned}$$

where $\text{Cost}(\ldots) = -\log P(\ldots)$.

# Speech Recognition Problems

- *Modeling:* how to describe accurately the relations between levels ⇒ *modeling errors*

- *Search:* how to find the best interpretation of the observations according to the given models ⇒ *search errors*

# Acoustic Modeling – Feature Selection I

- Short-time spectral analysis:

$$\log \left| \int g(\tau) x(t+\tau) e^{-i2\pi f \tau} \, d\tau \right|$$

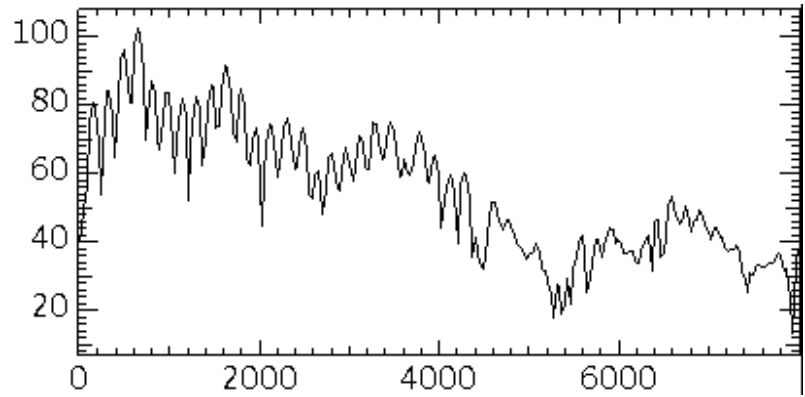

Short-time (25 msec. Hamming window) spectrum of /ae/ – Hz. vs. Db.

- Scale selection:

  – Cepstral smoothing

  – Parameter sampling (13 parameters)

# Acoustic Modeling – Feature Selection II [40, 38]

- Refinements

  - Time derivatives – 1st and 2nd order

  - non-Fourier analysis (e.g., Mel scale)

  - speaker/channel adaptation

    * mean cepstral subtraction
    * vocal tract normalization
    * linear transformations

- Result: 39 dimensional feature vector (13 cepstra, 13 delta cepstra, 13 delta-delta cepstra) every 10 milliseconds

# Acoustic Modeling – Stochastic Distributions [4, 61, 39, 5]

- Vector quantization – find codebook of prototypes

- Full covariance multivariate Gaussians:

$$P[\mathbf{y}] = \frac{1}{(2\pi)^{N/2}|\mathbf{S}|^{1/2}} e^{-\frac{1}{2}(\mathbf{y}^T - \mu^T)\mathbf{S}^{-1}(\mathbf{y}-\mu)}$$

- Diagonal covariance Gaussian mixtures

- Semi-continuous, tied mixtures

# Acoustic Modeling – Units and Training [61, 36]

- Units

  - Phonetic (*sub-word*) units – e.g., cat –> /k ae t/

  - Context-dependent units – $ae_{k,t}$

  - Multiple distributions (*states*) per phone – left, middle, right

- Training

  - *Given a segmentation*, training straight-forward

  - Obtain segmentation by *transcription*

  - Iterate until convergence

# Generating Lexicons – Two Steps

- Orthography → Phonemes

  "had" → /hh ae d/

  "your" → /y uw r/

  - complex, context-independent mapping
  - usually small number of alternatives
  - determined by spelling constraints; lexical "facts"
  - large online dictionaries available

- Phonemes → Phones

  /hh ae d y uw r/ → [hh ae dcl jh axr]    (60% prob)

  /hh ae d y uw r/ → [hh ae dcl d y axr]    (40% prob)

  - complex, context-dependent mapping
  - many possible alternatives
  - determined by phonological and phonetic constraints

M.Mohri-M.Riley-R.Sproat    Algorithms for Speech Recognition and Language Processing    PART II    63

# Decision Trees: Overview [9]

- **Description/Use:** Simple structure – binary tree of decisions, terminal nodes determine prediction *(cf. "Game of Twenty Questions")*. If dependent variable is categorical (e.g., `red, yellow, green`), called "classification tree", if continuous, called "regression tree".

- **Creation/Estimation:** Creating a binary decision tree for classification or regression involves three steps *(Breiman, et al)*:
  1. *Splitting Rules:* Which split to take at a node?
  2. *Stopping Rules:* When to declare a node terminal?
  3. *Node Assignment:* Which class/value to assign to a terminal node?

# 1. Decision Tree Splitting Rules

Which split to take at a node?

- **Candidate splits considered.**
    - *Binary cuts*: For **continuous** $-\infty \leq x < \infty$, consider splits of form:
    $$x \leq k \quad \text{vs.} \quad x > k, \quad \forall k.$$
    - *Binary partitions*: For **categorical** $x \in \{1, 2, ..., n\} = X$, consider splits of form:
    $$x \in A \quad \text{vs.} \quad x \in X - A, \quad \forall A \subset X.$$

# 1. Decision Tree Splitting Rules – Continued

- **Choosing best candidate split.**

  - *Method 1*: Choose $k$ (continuous) or $A$ (categorical) that minimizes estimated classification (regression) error after split.

  - *Method 2 (for classification)*: Choose $k$ or $A$ that minimizes estimated entropy after that split.

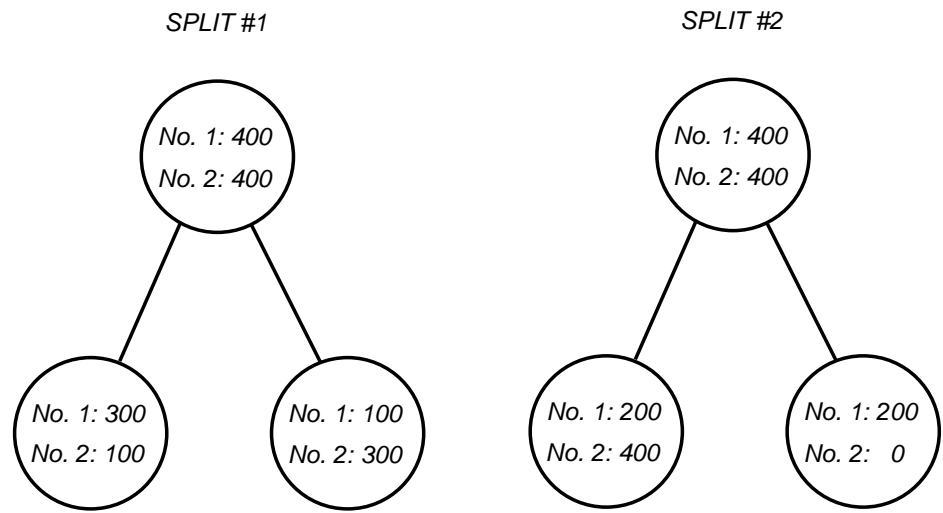

# 2. Decision Tree Stopping Rules

When to declare a node terminal?

- Strategy (*Cost-Complexity pruning*):
  1. Grow over-large tree.
  2. Form sequence of subtrees, $T_0, ..., T_n$ ranging from full tree to just the root node.
  3. Estimate "honest" error rate for each subtree.
  4. Choose tree size with mininum "honest" error rate.

- To form sequence of subtrees, vary $\alpha$ from 0 (for full tree) to $\infty$ (for just root node) in:
$$\min_T \left[ R(T) + \alpha |T| \right].$$

- To estimate "honest" error rate, test on data different from training data, e.g., grow tree on 9/10 of available data and test on 1/10 of data repeating 10 times and averaging (*cross-validation*).

# End of Declarative Sentence Prediction: Pruning Sequence

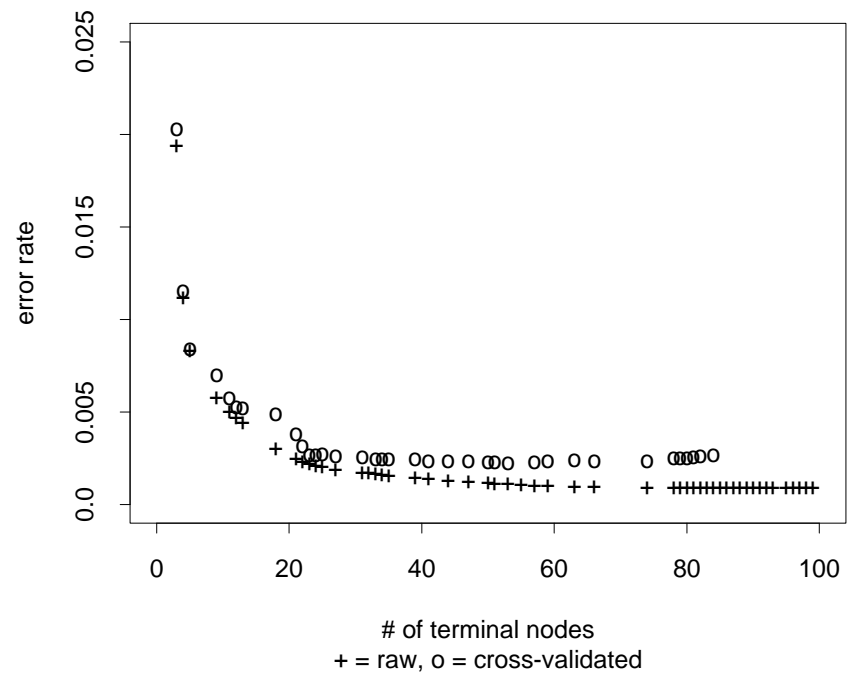

# 3. Decision Tree Node Assignment

Which class/value to assign to a terminal node?

- *Plurality vote*: Choose most frequent class at that node for classification; choose mean value for regression.

# End-of-Declarative-Sentence Prediction: Features [65]

- Prob[word with "." occurs at end of sentence]

- Prob[word after "." occurs at beginning of sentence]

- Length of word with "."

- Length of word after "."

- Case of word with ".": Upper, Lower, Cap, Numbers

- Case of word after ".": Upper, Lower, Cap, Numbers

- Punctuation after "." (if any)

- Abbreviation class of word with ".": – e.g., month name, unit-of-measure, title, address name, etc.

# End of Declarative Sentence?

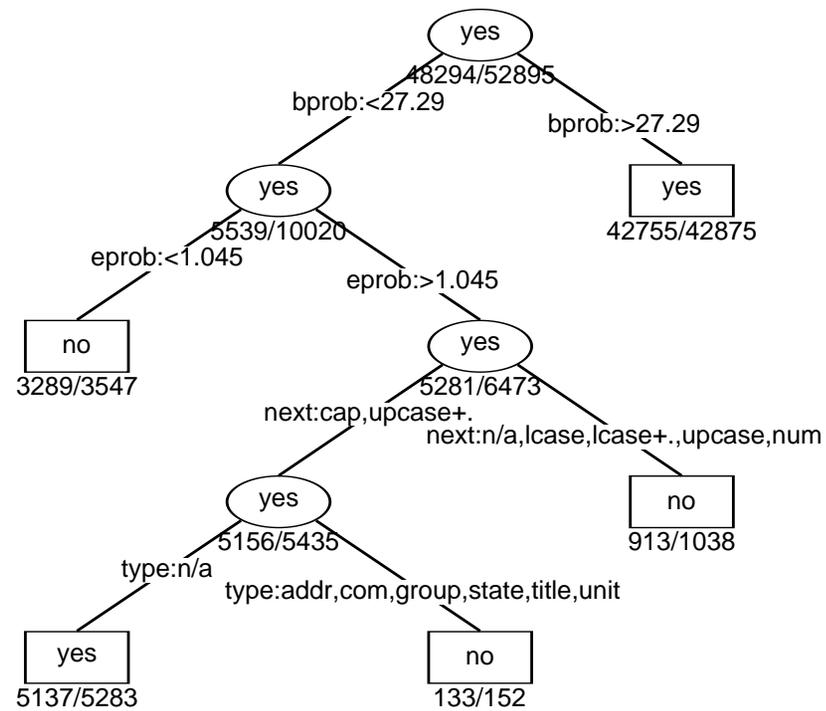

# Phoneme-to-Phone Alignment

| PHONEME | PHONE | WORD |
|---|---|---|
| p | p | purpose |
| er | er | |
| p | pcl | |
| - | p | |
| ax | ix | |
| s | s | |
| ae | ax | and |
| n | n | |
| d | - | |
| r | r | respect |
| ih | ix | |
| s | s | |
| p | pcl | |
| - | p | |
| eh | eh | |
| k | kcl | |
| t | t | |

# Phoneme-to-Phone Realization: Features [66, 10, 62]

- Phonemic Context:
  - Phoneme to predict
  - Three phonemes to left
  - Three phonemes to right

- Stress (0, 1, 2)

- Lexical Position:
  - Phoneme count from start of word
  - Phoneme count from end of word

# Phoneme-to-Phone Realization: Prediction Example

Tree splits for `/t/` in ``your pretty red'':

| PHONE | COUNT | SPLIT |
|-------|-------|-------|
| ix    | 182499 |      |
| n     | 87283  | cm0: vstp,ustp,vfri,ufri,vaff,uaff,nas |
| kcl+k | 38942  | cm0: vstp,ustp,vaff,uaff |
| tcl+t | 21852  | cp0: alv,pal |
| tcl+t | 11928  | cm0: ustp |
| tcl+t | 5918   | vm1: mono,rvow,wdi,ydi |
| dx    | 3639   | cm-1: ustp,rho,n/a |
| dx    | 2454   | rstr: n/a,no |

# Phoneme-to-Phone Realization: Network Example

Phonetic network for ``Don had your pretty...'':

| PHONEME | PHONE1 | PHONE2 | PHONE3 | CONTEXT |
|---|---|---|---|---|
| d | 0.91 d | | | |
| aa | 0.92 aa | | | |
| n | 0.98 n | | | |
| hh | 0.74 hh | 0.15 hv | | |
| ae | 0.73 ae | 0.19 eh | | |
| d | 0.51 dcl jh | 0.37 dcl d | | |
| y | 0.90 y | | | (if d→dcl d) |
| | 0.84 - | 0.16 y | | (if d→dcl jh) |
| uw | 0.48 axr | 0.29 er | | |
| r | 0.99 - | | | |
| p | 0.99 pcl p | | | |
| r | 0.99 r | | | |
| ih | 0.86 ih | | | |
| t | 0.73 dx | 0.11 tcl t | | |
| iy | 0.90 iy | | | |

# Acoustic Model Context Selection [92, 39]

- Statistical *regression* trees used to predict contexts based on distribution variance

- One tree per context-independent phone and state (left, middle, right)

- The trees were grown until the data criterion of 500 frames per distribution was met

- Trees pruned using cost-complexity pruning and cross-validation to select best contexts

- About 44000 context-dependent phone models

- About 16000 distributions

# N-Grams: Basics

- **'Chain Rule' and Joint/Conditional Probabilities:**

$$P[x_1 x_2 \ldots x_N] = P[x_N | x_1 \ldots x_{N-1}] P[x_{N-1} | x_1 \ldots x_{N-2}] \ldots P[x_2 | x_1] P[x_1]$$

where, e.g.,

$$P[x_N | x_1 \ldots x_{N-1}] = \frac{P[x_1 \ldots x_N]}{P[x_1 \ldots x_{N-1}]}$$

- **(First–Order) Markov assumption:**

$$P[x_k | x_1 \ldots x_{k-1}] = P[x_k | x_{k-1}] = \frac{P[x_{k-1} x_k]}{P[x_{k-1}]}$$

- **nth–Order Markov assumption:**

$$P[x_k | x_1 \ldots x_{k-1}] = P[x_k | x_{k-n} \ldots x_{k-1}] = \frac{P[x_{k-n} \ldots x_k]}{P[x_{k-n} \ldots x_{k-1}]}$$

# N-Grams: Maximum Likelihood Estimation

Let $N$ be total number of n-grams observed in a corpus and $c(x_1 \ldots x_n)$ be the number of times the n-gram $x_1 \ldots x_n$ occurred. Then

$$P[x_1 \ldots x_n] = \frac{c(x_1 \ldots x_n)}{N}$$

is the maximum likelihood estimate of that n-gram probability.

For conditional probabilities,

$$P[x_n | x_1 \ldots x_{n-1}] = \frac{c(x_1 \ldots x_n)}{c(x_1 \ldots x_{n-1})}.$$

is the maximum likelihood estimate.

With this method, an n-gram that does not occur in the corpus is assigned **zero** probability.

# N-Grams: Good-Turing-Katz Estimation [29, 16]

Let $n_r$ be the number of n-grams that occurred $r$ times. Then

$$P[x_1 \ldots x_n] = \frac{c^*(x_1 \ldots x_n)}{N}$$

is the Good-Turing estimate of that n-gram probability, where $c^*(x) = (c(x) + 1)\frac{n_{c(x)+1}}{n_{c(x)}}$.

For conditional probabilities,

$$P[x_n | x_1 \ldots x_{n-1}] = \frac{c^*(x_1 \ldots x_n)}{c(x_1 \ldots x_{n-1})}, \qquad c(x_1 \ldots x_n) > 0$$

is Katz's extension of the Good-Turing estimate.

With this method, an n-gram that does not occur in the corpus is assigned the backoff probability $P[x_n | x_1 \ldots x_{n-1}] = \alpha P[x_n | x_2 \ldots x_{n-1}]$, where $\alpha$ is a normalizing constant.

# Finite-State Modeling [57]

*Our view of recognition cascades*: represent mappings between levels, observation sequences and language uniformly with *weighted* finite-state machines:

- Probabilistic mapping $P(\mathbf{x}|\mathbf{y})$: *weighted finite-state transducer*. Example — word pronunciation transducer:

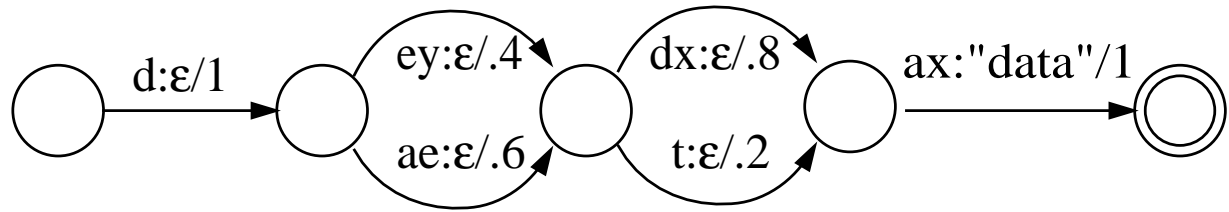

- Language model $P(\mathbf{w})$: *weighted finite-state acceptor*

# Example of Recognition Cascade

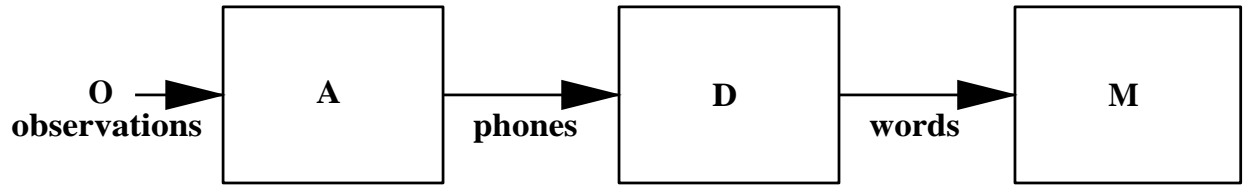

- Recognition from observations **o** by composition:
  - *Observations:* $O(\mathbf{s}, \mathbf{s}) = \begin{cases} 1 & \text{if } \mathbf{s} = \mathbf{o} \\ 0 & \text{otherwise} \end{cases}$
  - *Acoustic-phone transducer:* $A(\mathbf{a}, \mathbf{p}) = P(\mathbf{a}|\mathbf{p})$
  - *Pronunciation dictionary:* $D(\mathbf{p}, \mathbf{w}) = P(\mathbf{p}|\mathbf{w})$
  - *Language model:* $M(\mathbf{w}, \mathbf{w}) = P(\mathbf{w})$
- *Recognition:* $\hat{\mathbf{w}} = \underset{\mathbf{w}}{\mathrm{argmax}}(O \circ A \circ D \circ M)(\mathbf{o}, \mathbf{w})$

# Speech Models as Weighted Automata

- Quantized observations:

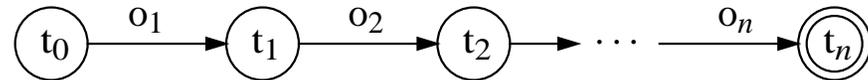

- Phone model $A_\pi$ : observations $\to$ phones

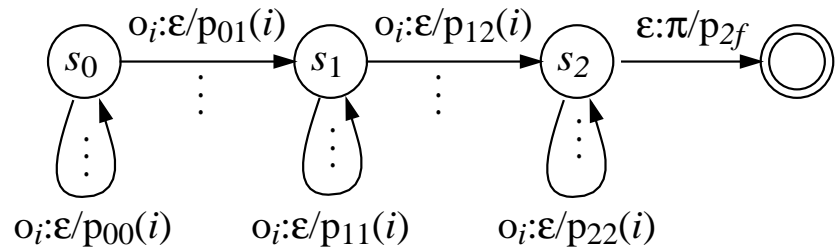

Acoustic transducer: $A = \left(\sum_\pi A_\pi\right)^*$

- Word pronunciations $D_{\text{data}}$ : phones $\to$ words

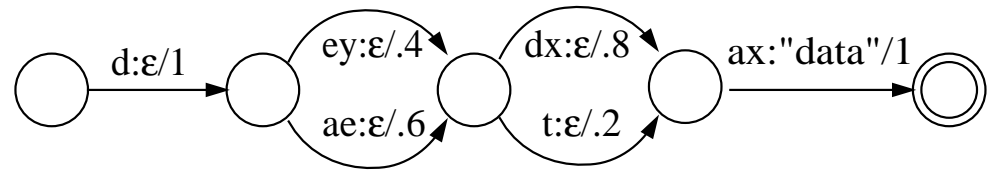

Dictionary: $D = \left(\sum_w D_w\right)^*$

# Example: Phone Lattice $O \circ A$

- *Lattices:* Weighted acyclic graphs representing possible interpretations of an utterance as sequences of units at a given level of representation (phones, syllables, words,...)

- Example: result of composing observation sequence for *hostile battle* with acoustic model:

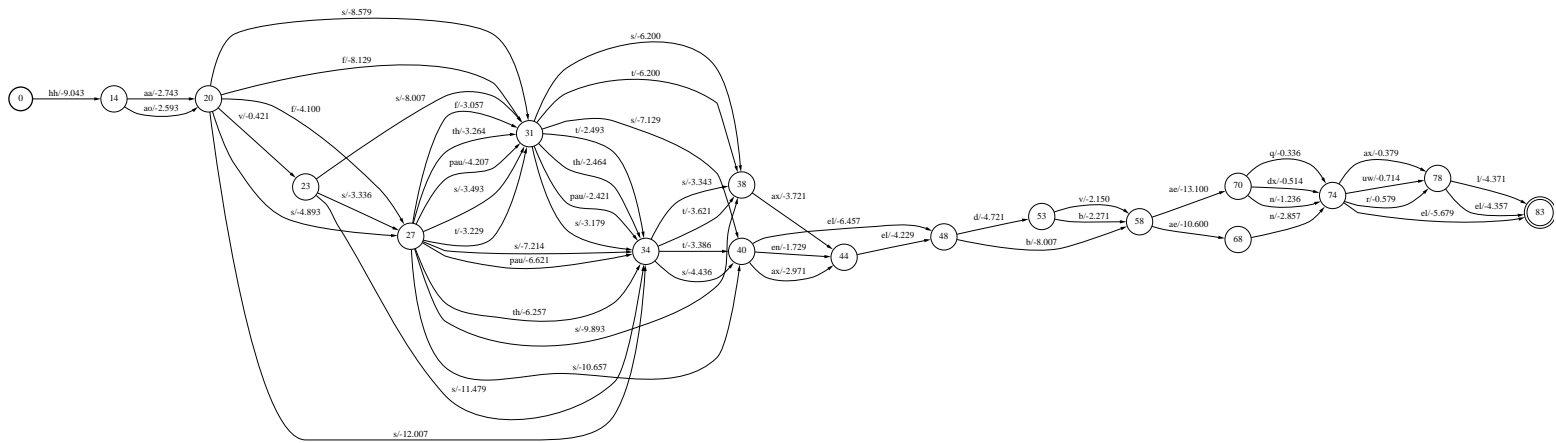

# Sample Pronunciation Dictionary $D$

Dictionary with *hostile*, *battle* and *bottle* as a weighted transducer:

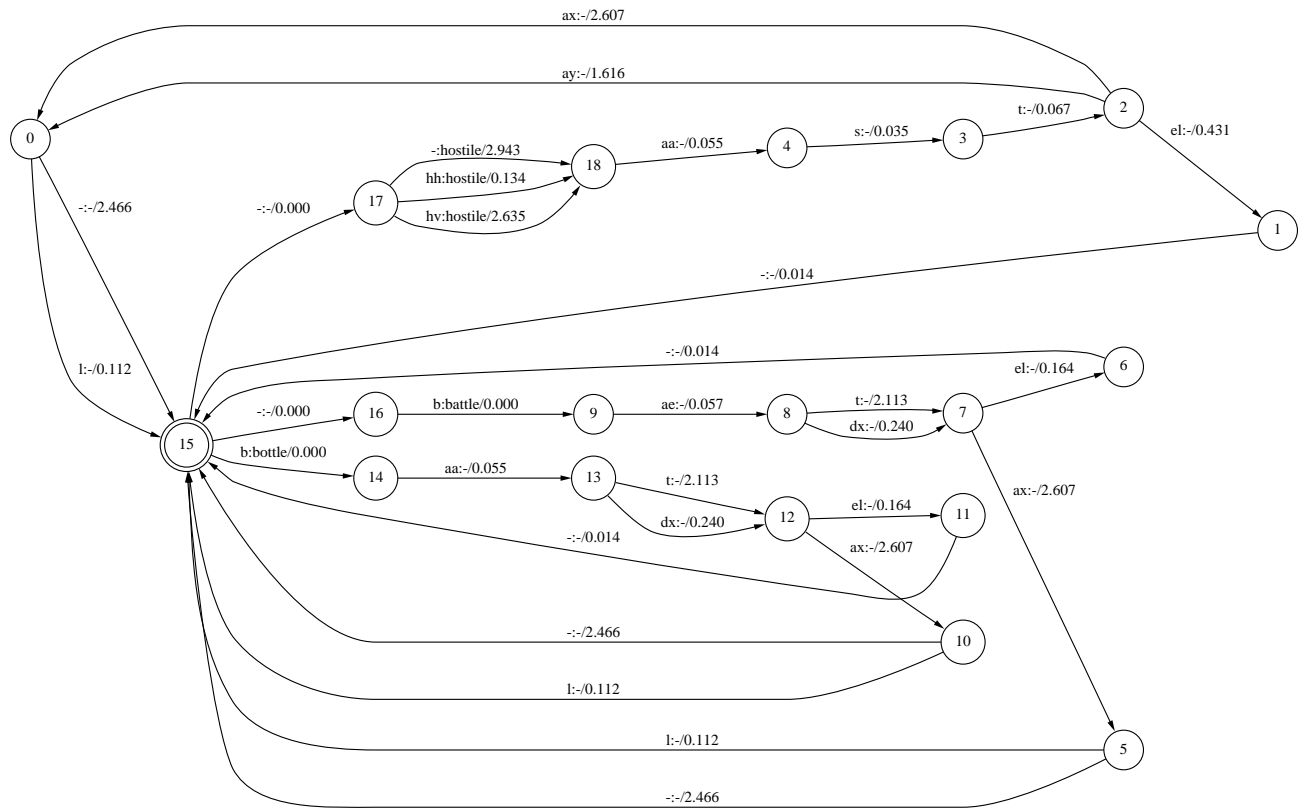

# Sample Language Model $M$

Simplified language model as a weighted acceptor:

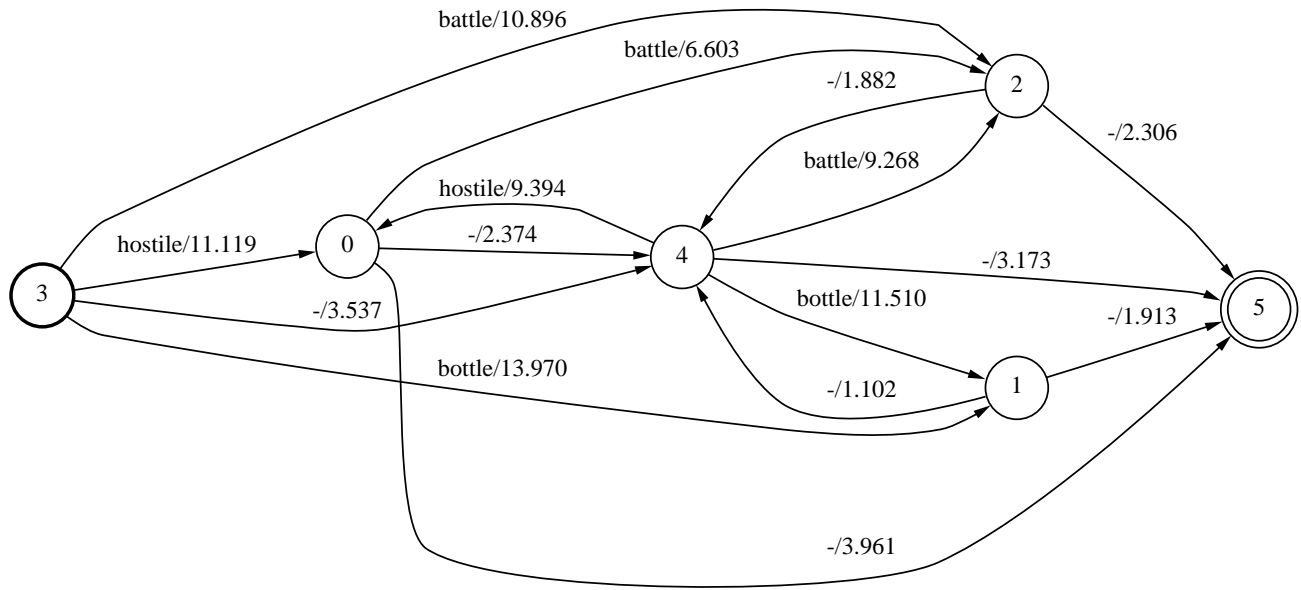

# Recognition by Composition

- *From phones to words:* compose dictionary with phone lattice to yield word lattice with combined acoustic and pronunciation costs:

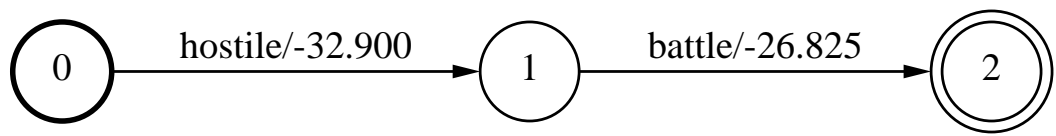

- *Applying language model:* Compose word lattice with language model to obtain word lattice with combined acoustic, pronunciation and language model costs:

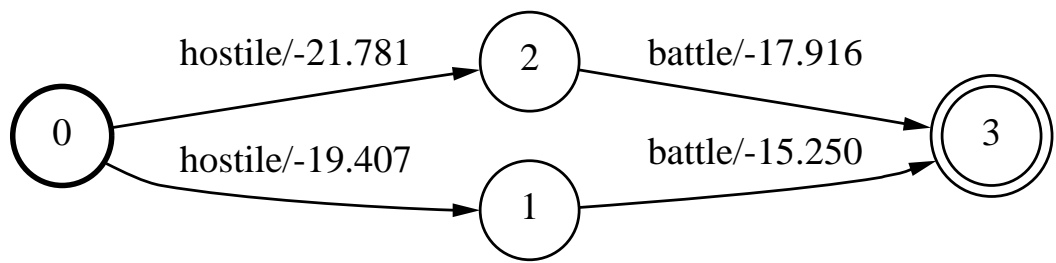

# Context-Dependency Examples

- **Context-dependent phone models:** Maps from CI units to CD units. Example: $ae/b\_\_d \rightarrow ae_{b,d}$

- **Context-dependent allophonic rules:** Maps from baseforms to detailed phones. Example: $t/V'\_\_V \rightarrow dx$

- **Difficulty:** Cross-word contexts – where several words enter and leave a state in the grammar, substitution does not apply.

# Context-Dependency Transducers

Example — triphonic context transducer for two symbols $x$ and $y$.

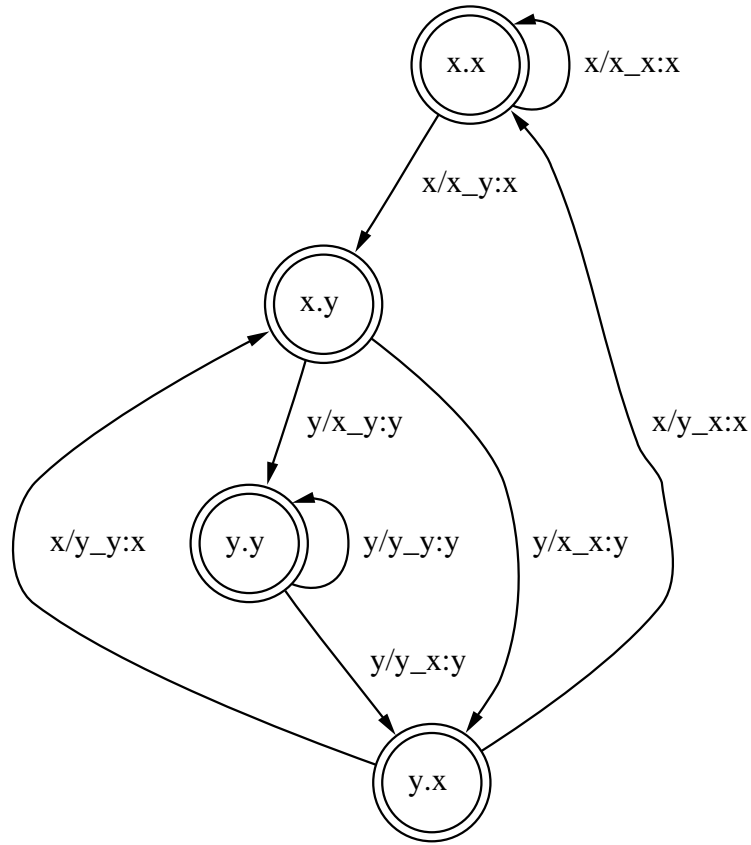

# Generalized State Machines

All of the above networks have *bounded context* and thus can be represented as *generalized state machines*. A generalized state machine $M$:

- **Supports these operations:**
  - $M.start$ – returns start state
  - $M.final(state)$ – returns 1 if final, 0 if non-final state
  - $M.arcs(state)$ – returns transitions $(a_1, a_2, \ldots, a_N)$ leaving $state$, where $a_i = (ilabel, olabel, weight, nextstate)$

- **Does *not* necessarily support:**
  - providing the number of states
  - expanding states that have not been already *discovered*

# On-Demand Composition [69, 53]

Create generalized state machine $C$ for composition $A \circ B$.

- $C.start := (A.start, B.start)$

- $C.final((s1, s2)) := A.final(s1) \wedge B.final(s2)$

- $C.arcs((s1, s2)) := Merge(A.arcs(s1), B.arcs(s2))$

Merged arcs defined as:

$$(l1, l3, x + y, (ns1, ns2)) \in Merge(A.arcs(s1), B.arcs(s2))$$

iff

$$(l1, l2, x, ns1) \in A.arcs(s1) \text{ and } (l2, l3, y, ns2) \in B.arcs(s2)$$

# State Caching

Create generalized state machine $B$ for input machine $A$.

- $B.start := A.start$
- $B.final(state) := A.final(state)$
- $B.arcs(state) := A.arcs(state)$

Cache Disciplines:

- Expand each state of A exactly once, i.e. always save in cache (memoize).
- Cache, but forget 'old' states using a least-recently used criterion.
- Use instructions (ref counts) from user (decoder) to save and forget.

# On Demand Composition – Results

ATIS Task - class-based trigram grammar, full cross-word triphonic context-dependency.

|                | states              | arcs              |
|----------------|---------------------|-------------------|
| context        | 762                 | 40386             |
| lexicon        | 3150                | 4816              |
| grammar        | 48758               | 359532            |
| full expansion | $\sim 1.6 \times 10^6$ | $5.1 \times \sim 10^6$ |

For the same recognition accuracy as with a static, fully expanded network, on-demand composition expands just 1.6% of the total number of arcs.

# Determinization in Large Vocabulary Recognition

- For large vocabularies, 'string' lexicons are very non-deterministic

- Determinizing the lexicon solves this problem, but can introduce non-coassessible states during its composition with the grammar

- Alternate Solutions:
  - Off-line compose, determinize, and minimize:

  $$Lexicon \circ Grammar$$

  - Pre-tabulate non-coassessible states in the composition of:

  $$Det(Lexicon) \circ Grammar$$

# Search in Recognition Cascades

- Reminder: $Cost \equiv -\log probability$

- Example recognition problem: $\hat{\mathbf{w}} = \operatorname*{argmax}_{\mathbf{w}} (O \circ A \circ D \circ M)(\mathbf{o}, \mathbf{w})$

- *Viterbi search*: approximate $\hat{\mathbf{w}}$ by the output word sequence for the lowest-cost path from the start state to a final state in $O \circ A \circ D \circ M$ — ignores summing over multiple paths with same output:

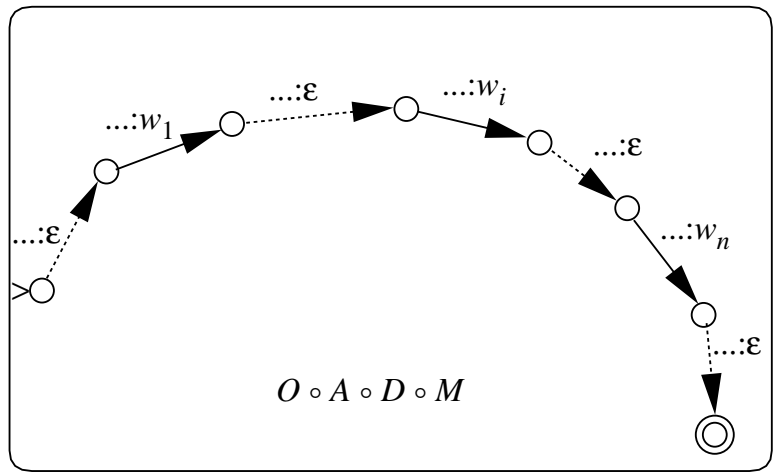

$O \circ A \circ D \circ M$

- Composition preserves acyclicity, $O$ is acyclic $\Rightarrow$ acyclic search graph

# Single-source Shortest Path Algorithms [83]

- Meta-algorithm:

  $Q \leftarrow \{s_0\}; \forall s, Cost(s) \leftarrow \infty$

  **While** $Q$ not empty, $s \leftarrow \text{DEQUEUE}(Q)$

  **For** each $s' \in Adj[s]$ such that $Cost(s') > Cost(s) + cost(s, s')$

  $Cost(s') \leftarrow Cost(s) + cost(s, s')$

  $\text{ENQUEUE}(Q, s)$

- Specific algorithms:

| Name | Queue type | Cycles | Neg. Weights | Complexity |
|---|---|---|---|---|
| acyclic | topological | no | yes | $O(|V| + |E|)$ |
| Dijkstra | best-first | yes | no | $O(|E| \log |V|)$ |
| Bellman-Ford | FIFO | yes | yes | $O(|V| \cdot |E|)$ |

# The Search Problem

- *Obvious first approach*: use an appropriate single-source shortest-path algorithm

- *Problem:* impractical to visit all states, can we do better?
  - *Admissible* methods: guarantee finding best path, but reorder search to avoid exploring provably bad regions
  - *Non-admissible* methods: may fail to find best path, but may need to explore much less of the graph

- Current practical approaches:
  - Heuristic cost functions
  - Beam search
  - Multipass search
  - Rescoring

# Heuristic Cost Function — A* Search [4, 56, 17]

- States in search ordered by

$$\text{cost-so-far}(s) + \text{lower-bound-to-complete}(s)$$

- With a tight bound, states not on good paths are not explored

- With a loose lower bound no better than Dijkstra's algorithm

- Where to find a tight bound?

  – Full search of a composition of smaller automata (homomorphic automata with lower-bounding costs?)

  – Non-admissible A* variants: use averaged estimate of cost-to-complete, not a lower-bound

# Beam Search [35]

- Only explore states with costs within a *beam* (threshold) of the cost of the best *comparable* state

- Non-admissible

- *Comparable states* ≡ states corresponding to (approximately) the same observations

- Synchronous (Viterbi) search: explore composition states in chronological observation order

- Problem with synchronous beam search: too local, some observation subsequences are unreliable and may locally put the best overall path outside the beam

## Beam-Search Tradeoffs [68]

*Word lattice:* result of composing observation sequence, level transducers and language model.

| Beam | Word lattice error rate | Median number of edges |
|------|-------------------------|------------------------|
| 4    | 7.3%                    | 86.5                   |
| 6    | 5.4%                    | 244.5                  |
| 8    | 4.4%                    | 827                    |
| 10   | 4.1%                    | 3520                   |
| 12   | 4.0%                    | 13813.5                |

# Multipass Search [52, 3, 68]

- Use a succession of binary compositions instead of a single $n$-way composition — combinable with other methods

- *Prune*: Use two-pass variant of composition to remove states not in any path close enough to the best

- Pruned intermediate lattices are smaller, lower number of state pairings considered

- *Approximate*: use simpler models (context-independent phone models, low-order language models)

- *Rescore*...

# Rescoring

Most successful approach in practice:

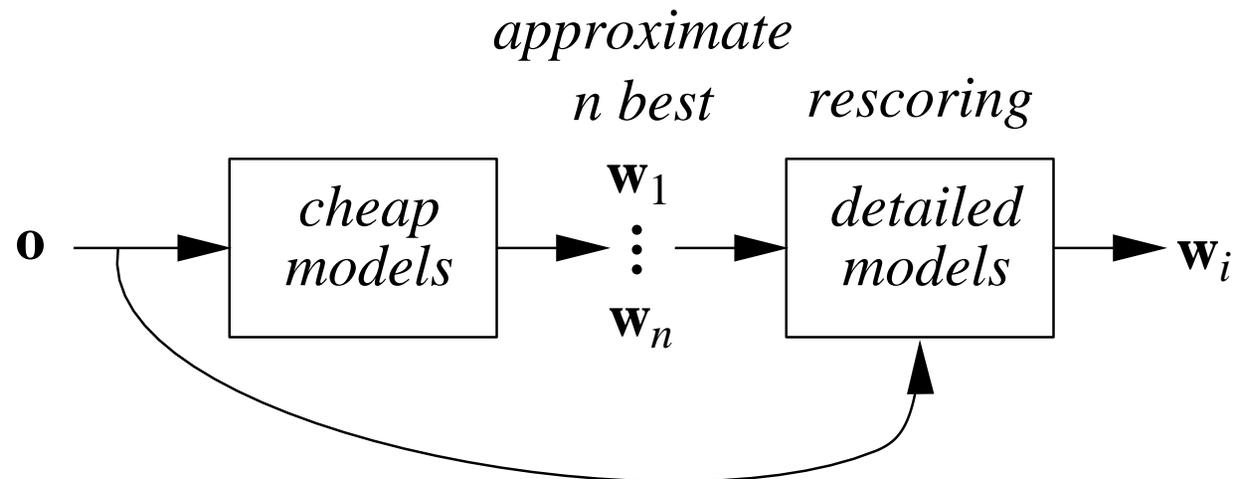

- Small pruned result built by composing approximate models

- Composition with full models, observations

- Find lowest-cost path

# PART III
# Finite State Methods in Language Processing

Richard Sproat

Speech Synthesis Research Department

Bell Laboratories, Lucent Technologies

`rws@bell-labs.com`

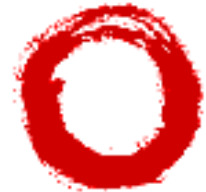

# Overview

- Text-analysis for Text-to-Speech (TTS) Synthesis

  – A rich domain with lots of linguistic problems

  – Probably the least familiar application of NLP technologies

- Syntactic analysis

- Some thoughts on text indexation

# The Nature of the TTS Problem

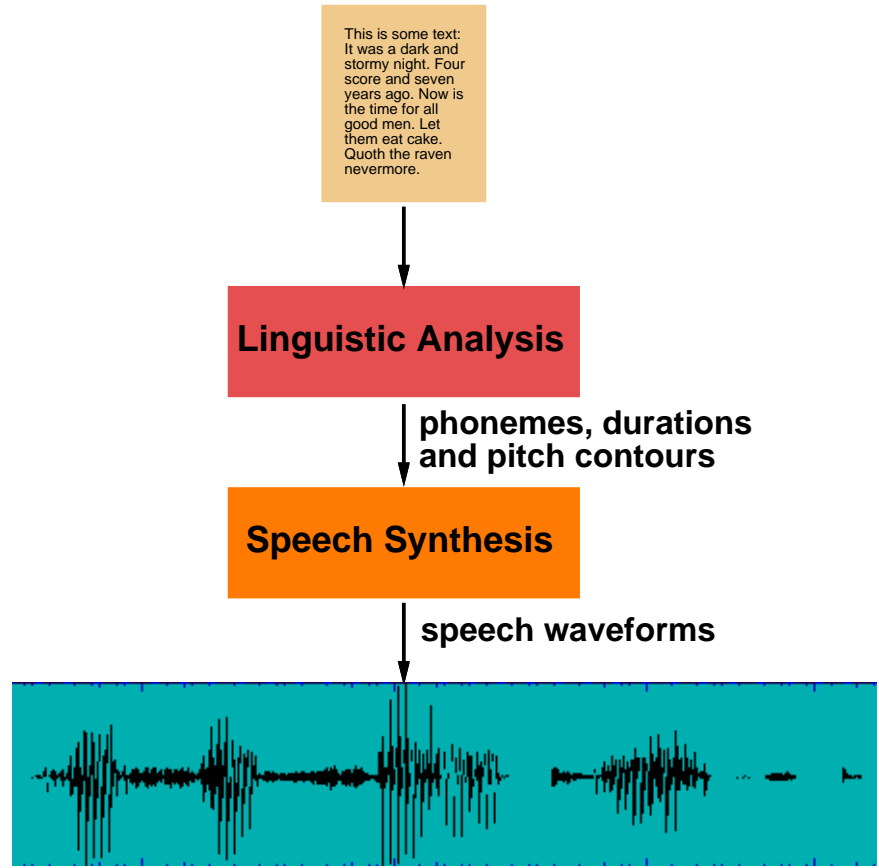

# From Text to Linguistic Representation

老鼠吃油。

'The rat is eating the oil'

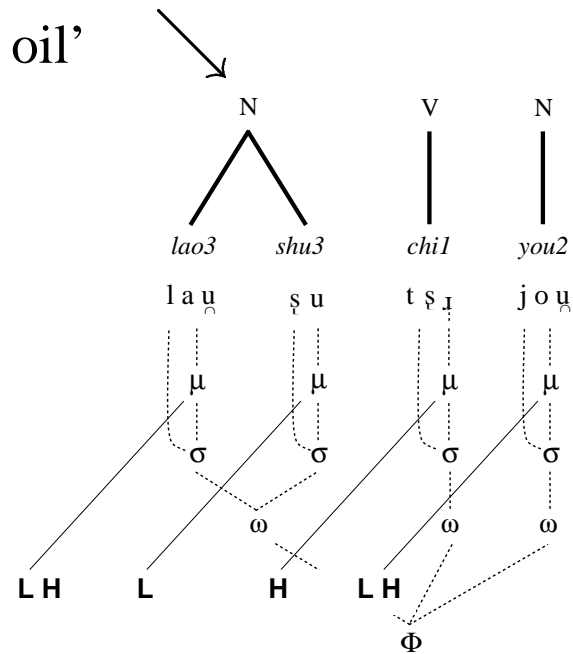

# Russian Percentages: The Problem

How do you say '%' in Russian?

| | | *Adjectival forms when modifying nouns* |
|---|---|---|
| 20% скидка | ⇒ | двадцат\[и\]-процентн\[ая\] скидка |
| '20% discount' | | *dvadcat\[i\]-procent\[naja\] skidka* |
| с 20% раствором | ⇒ | с двадцат\[и\]-процент\[ным\] раствором |
| 'with 20% solution' | | *s dvadcat\[i\]-procent\[nym\] rastvorom* |
| | | *Nominal forms otherwise* |
| 21% | ⇒ | двадцать один процент |
| | | *dvadcat' odin procent* |
| 23% | ⇒ | двадцать три процент\[а\] |
| | | *dvadcat' tri procent\[a\]* |
| 20% | ⇒ | двадцать процент\[ов\] |
| | | *dvadcat' procent\[ov\]* |
| с 20% | ⇒ | с двадцать\[ю\] процент\[ами\] |
| 'with 20%' | | *s dvadcat'\[ju\] procent\[ami\]* |

# Text Analysis Problems

- Segment text into words.
- Segment text into sentences, checking for and expanding abbreviations :

*St. Louis is in Missouri.*

- Expand numbers
- Lexical and morphological analysis
- Word pronunciation
  - Homograph disambiguation
- Phrasing
- Accentuation

# Desiderata for a Model of Text Analysis for TTS

- Delay decisions until have enough information to make them

- Possibly *weight* various alternatives

Weighted Finite-State Transducers offer an attractive computational model

# Overall Architectural Matters

Example: word pronunciation in Russian

- Text form: костра<kostra> (bonfire+genitive.singular)

- Morphological analysis:
  кост'Ёр{noun}{masc}{inan}+'a{sg}{gen}

- Pronunciation: /kʌstr'a/

- Minimal Morphologically-Motivated Annotation (MMA): костр'а

(Sproat, 1996)

# Overall Architectural Matters

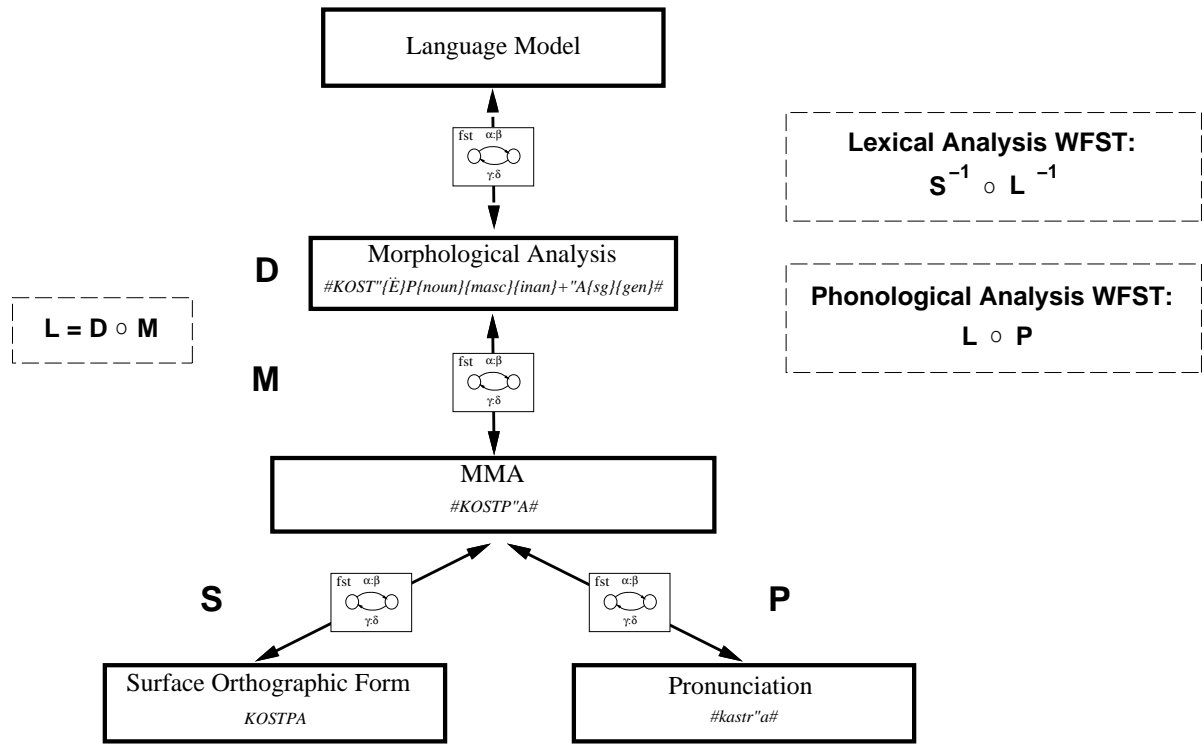

# Orthography → Lexical Representation

## A Closer Look

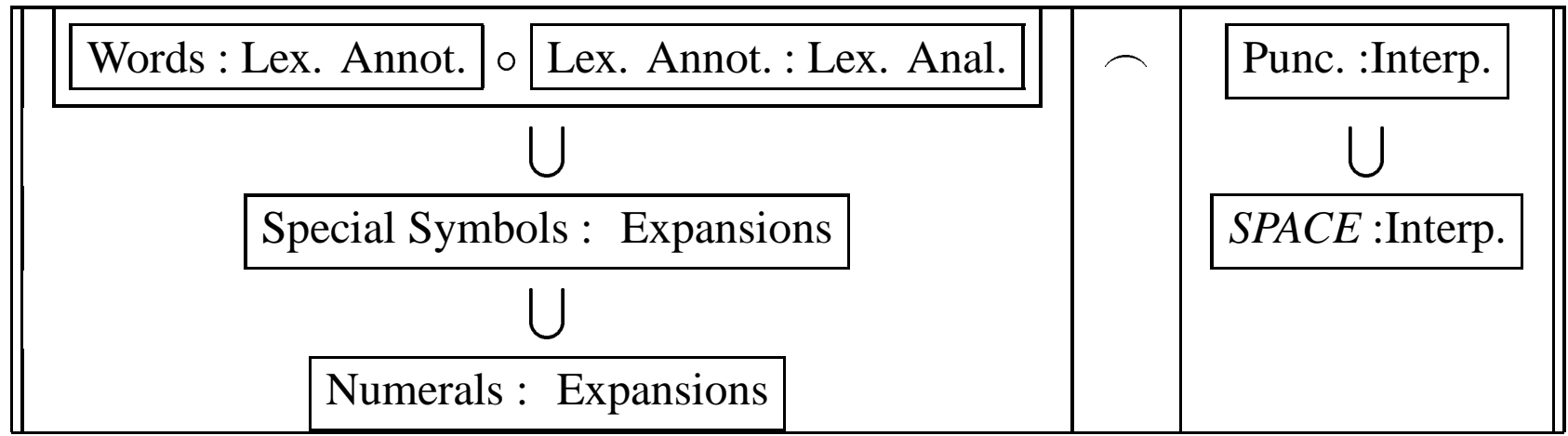

*SPACE*:  white space in German, Spanish, Russian ...

  $\epsilon$ in Japanese, Chinese ...

# Chinese Word Segmentation

| | | | | |
|---|---|---|---|---|
| 了 | → | 了$_1$ asp$_{4.68}$ | *le0* | PERF |
| 了解 | → | 了$_2$解$_1$ vb$_{8.11}$ | *liao3jie3* | understand |
| 大 | → | 大$_1$ vb$_{5.56}$ | *da4* | big |
| 大街 | → | 大$_1$街 nc$_{11.45}$ | *da4jie1* | avenue |
| 不 | → | 不$_2$ adv$_{4.58}$ | *bu4* | not |
| 在 | → | 在 vb$_{4.45}$ | *zai4* | at |
| 忘 | → | 忘 vb$_{11.77}$ | *wang4* | forget |
| 忘不了 | → | 忘vb++不$_2$了$_2$ npot$_{12.23}$ | *wang4+bu4liao3* | unable to forget |
| 我 | → | 我 np$_{4.88}$ | *wo3* | I |
| 放 | → | 放 vb$_{8.05}$ | *fang4* | place |
| 放大 | → | 放大$_1$ vb$_{10.70}$ | *fang4da4* | enlarge |
| 哪裡 | → | 哪$_1$裡 nc$_{11.02}$ | *na3li3* | where |
| 街 | → | 街 nc$_{10.35}$ | *jie1* | avenue |
| 解放 | → | 解$_1$放 nc$_{10.92}$ | *jie3fang4* | liberation |
| 解放大 | → | 解$_3$放大$_1$ urnp$_{42.23}$ | *xie4 fang4da4* | NAME |

## Chinese Word Segmentation

Space = $\epsilon$ : #

L = $Space \frown (Dictionary \frown (Space \cup Punc))^+$

BestPath(我忘不了解放大街在哪裡。L) =
我pro$_{4.88}$#忘vb+不了$_2$npot$_{12.23}$# 解$_1$放nc$_{10.92}$大$_1$街nc$_{11.45}$ . . .
'I couldn't forget where Liberation Avenue is.'

## Numeral Expansion

$234 \quad \circ \quad \boxed{\text{Factorization}} \quad \Longrightarrow \quad 2 \cdot 10^2 + 3 \cdot 10^1 + 4$

$\quad \quad \quad \circ \quad \boxed{\text{DecadeFlop}} \quad \Longrightarrow \quad 2 \cdot 10^2 + 4 + 3 \cdot 10^1$

$\quad \quad \quad \circ \quad \boxed{\text{NumberLexicon}}^* \quad \Downarrow$

*zwei+hundert+vier+und+dreißig*

# Numeral Expansion

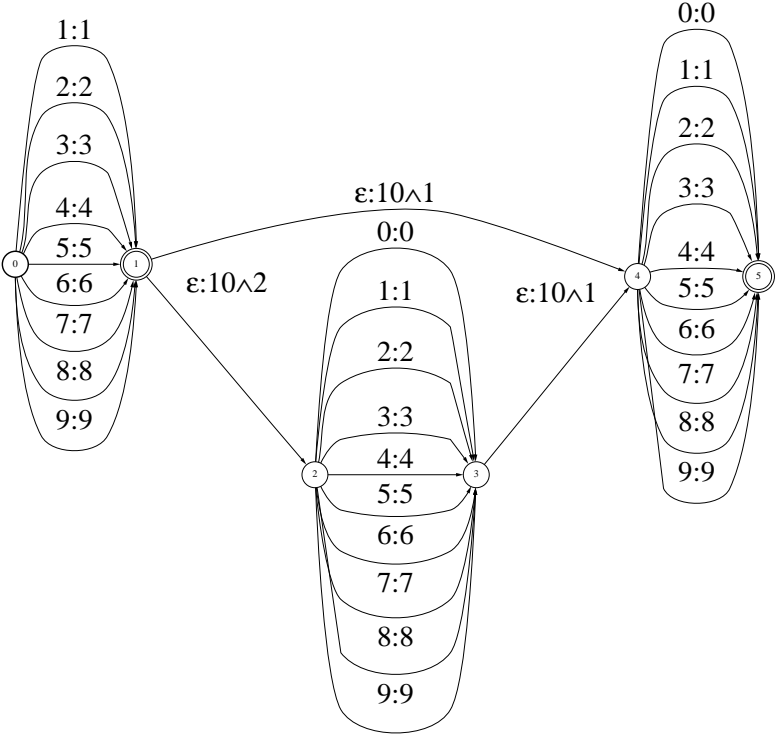

# German Numeral Lexicon

| | | |
|---|---|---|
| /{1} | : | ('eins{num}({masc}|{neut}){sg}{##})/ |
| /{2} | : | (zw'ei{num}{##})/ |
| /{3} | : | (dr'ei{num}{##})/ |
| ⋮ | | |
| /({0}{+++}{1}{10∧1}) | : | (z'ehn{num}{##})/ |
| /({1}{+++}{1}{10∧1}) | : | ('elf{num}{##})/ |
| /({2}{+++}{1}{10∧1}) | : | (zw'ölf{num}{##})/ |
| /({3}{+++}{1}{10∧1}) | : | (dr'ei{++}zehn{num}{##})/ |
| ⋮ | | |
| /({2}{10∧1}) | : | (zw'an{++}zig{num}{##})/ |
| /({3}{10∧1}) | : | (dr'ei{++}ßig{num}{##})/ |
| ⋮ | | |
| /({10∧2}) | : | (h'undert{num}{##})/ |
| /({10∧3}) | : | (t'ausend{num}{neut}{##})/ |

# Morphology: Paradigmatic Specifications

Paradigm    {A1}

                # starke Flektion (z.B. nach unbestimmtem Artikel)

Suffix      {++}er    {sg}{masc}{nom}

Suffix      {++}en    {sg}{masc}({gen}|{dat}|{acc})

Suffix      {++}e     {sg}{femi}({nom}|{acc})

Suffix      {++}en    {sg}({femi}|{neut})({gen}|{dat})

Suffix      {++}es    {sg}{neut}({nom}|{acc})

Suffix      {++}e     {pl}({nom}|{acc})

Suffix      {++}er    {pl}{gen}

Suffix      {++}en    {pl}{dat}

# Morphology: Paradigmatic Specifications

##### Possessiva ("mein, euer")

| | | |
|---|---|---|
| Paradigm | {A6} | |
| Suffix | {++}{Eps} | {sg}({masc}|{neut}){nom} |
| Suffix | {++}e | {sg}{femi}{nom} |
| Suffix | {++}es | {sg}({masc}|{neut}){gen} |
| Suffix | {++}er | {sg}{femi}({gen}|{dat}) |
| Suffix | {++}em | {sg}({masc}|{neut}){dat} |
| Suffix | {++}en | {sg}{masc}{acc} |
| Suffix | {++}{Eps} | {sg}{neut}{acc} |
| Suffix | {++}e | {pl}({nom}|{acc}) |
| Suffix | {++}er | {pl}{gen} |
| Suffix | {++}en | {pl}{dat} |

# Morphology: Paradigmatic Specifications

/{A1}   :   ('aal{++}glatt{adj})/

/{A1}   :   ('ab{++}änder{++}lich{adj}{umlt})/

/{A1}   :   ('ab{++}artig{adj})/

/{A1}   :   ('ab{++}bau{++}würdig{adj}{umlt})/

⋮

/{A6}   :   (d'ein{adj})/

/{A6}   :   ('euer{adj})/

/{A6}   :   ('ihr{adj})/

/{A6}   :   ('Ihr{adj})/

/{A6}   :   (m'ein{adj})/

/{A6}   :   (s'ein{adj})/

/{A6}   :   ('unser{adj})/

# Morphology: Paradigmatic Specifications

Project(({A6}⌢Endings) ∘ (({A6}:Stems)⌢Id(Σ*))) ⇒

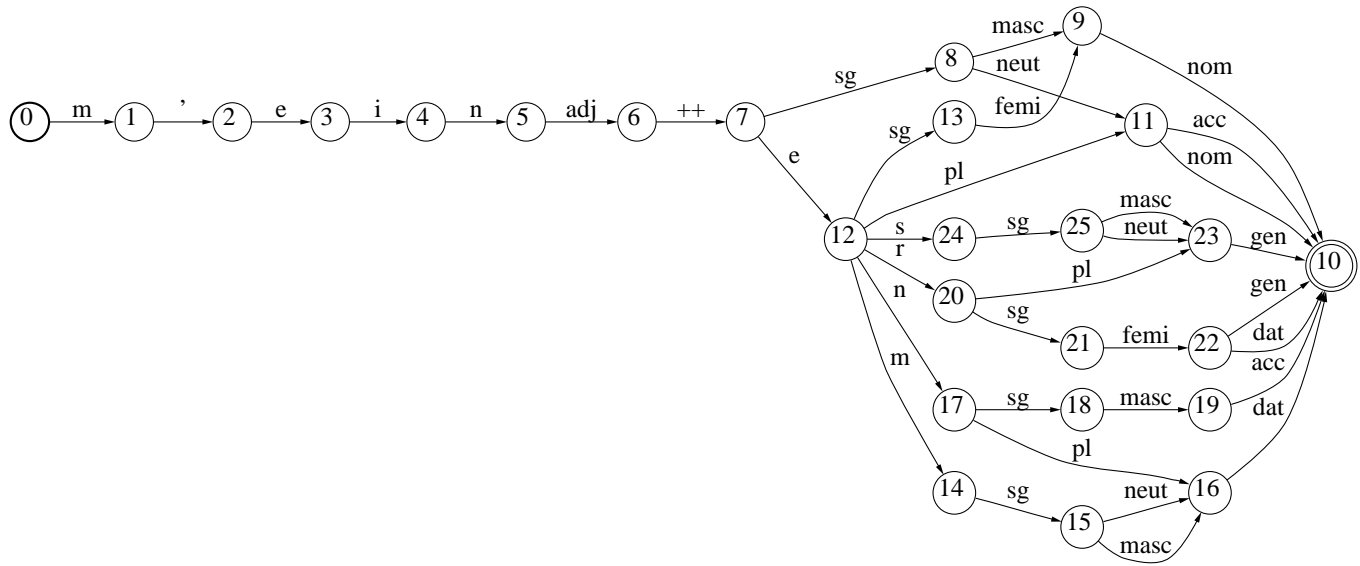

# Morphology: Finite-State Grammar

| | | |
|---|---|---|
| START   | PREFIX | {Eps} |
| PREFIX  | STEM   | {Eps} |
| PREFIX  | STEM   | t"ele{++}<1.0> |
| | ⋮ | |
| STEM    | SUFFIX | 'abend |
| STEM    | SUFFIX | 'abenteuer |
| | ⋮ | |
| SUFFIX  | PREFIX | {++}<1.0> |
| SUFFIX  | FUGE   | {Eps}<1.0> |
| SUFFIX  | WORD   | {Eps}<2.0> |
| | ⋮ | |

# Morphology: Finite-State Grammar

| | | |
|---|---|---|
| FUGE | SECOND | {++}<1.5> |
| FUGE | SECOND | {++}s{++}<1.5> |
| | ⋮ | |
| SECOND | PREFIX | {Eps}<1.0> |
| SECOND | STEM | {Eps}<2.0> |
| SECOND | WORD | {Eps}<2.0> |
| | ⋮ | |
| WORD | | |

# Morphology: Finite-State Grammar

Unanständigkeitsunterstellung

'allegation of indecency'

"un{++}"an{++}st'änd{++}ig{++}keit{++}s{++}unter{++}st'ell{++}ung

# Rewrite Rule Compilation

## Context-dependent rewrite rules

**General form**:
$$\phi \to \psi / \lambda \underline{\phantom{xx}} \rho$$

$\phi, \psi, \lambda, \rho$ regular expressions.

Constraint: $\psi$ cannot be rewritten but can be used as a context

**Example**:
$$a \to b / c \underline{\phantom{xx}} b$$

(Johnson, 1972; Kaplan & Kay, 1994; Karttunen, 1995; Mohri & Sproat, 1996)

# Example

$$a \to b/c\_\_b$$

$$w = cab$$

# Example

### Input:

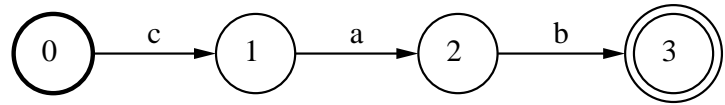

### After *r*:

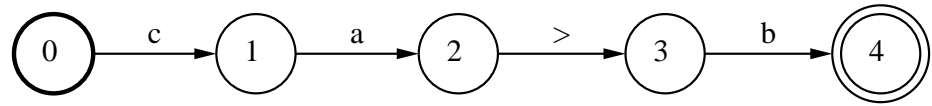

### After *f*:

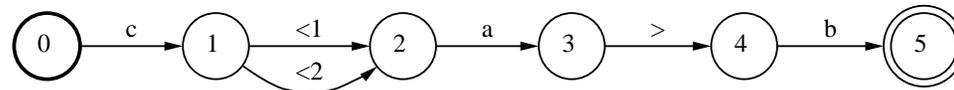

# Example

**After *replace*:**

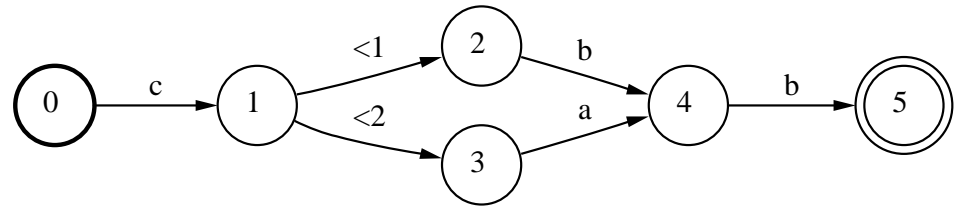

**After $l_1$:**

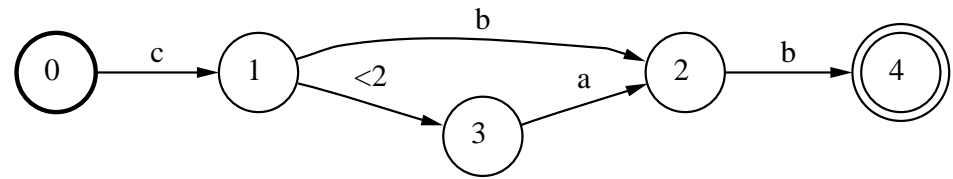

**After $l_2$:**

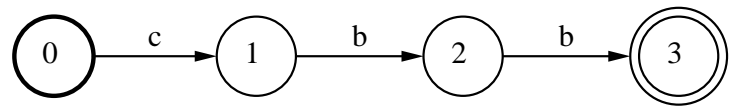

# Rewrite Rule Compilation

- Principle

  – Based on the use of marking transducers

  – Brackets inserted only where needed

- Efficiency

  – 3 determinizations + additional linear time work

  – Smaller number of compositions

# Rule Compilation Method

$$r \circ f \circ replace \circ l_1 \circ l_2$$

$r:$ $\quad \Sigma^* \rho \rightarrow \Sigma^* > \rho$

$f:$ $\quad (\Sigma \cup \{>\})^* \phi > \rightarrow (\Sigma \cup \{>\})^* \{<_1, <_2\} \phi >$

$replace:$ $\quad <_1 \phi > \rightarrow <_1 \psi$

$l_1:$ $\quad \Sigma^* \lambda <_1 \rightarrow \Sigma^* \lambda$

$l_2:$ $\quad \Sigma^* \overline{\lambda} <_2 \rightarrow \Sigma^* \overline{\lambda}$

# Marking Transducers

**Proposition** Let $\alpha$ be a deterministic automaton representing $\Sigma^*\beta$, then the transducer $\tau$ post-marks occurrences of $\beta$ by #.

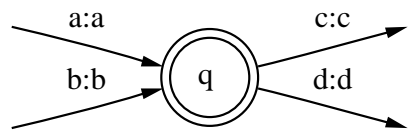

Final state $q$ with entering and leaving transitions of $Id(\alpha)$.

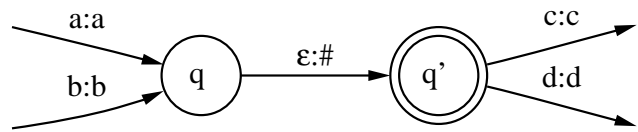

States and transitions after modifications, transducer $\tau$.

# Marker of Type 2

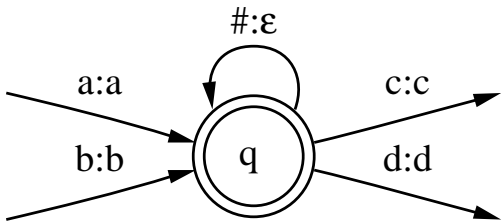

## The Transducers as Expressions using *Marker*

$$r = [reverse(Marker(\Sigma^* reverse(\rho), 1, \{>\}, \emptyset))]$$

$$f = [reverse(Marker((\Sigma \cup \{>\})^* reverse(\phi_> >), 1, \{<_1, <_2\}, \emptyset))]$$

$$l_1 = [Marker(\Sigma^* \lambda, 2, \emptyset, \{<_1\})]_{<_2:<_2}$$

$$l_2 = [Marker(\Sigma^* \lambda, 3, \emptyset, \{<_2\})]$$

# Example: $r$ for rule $a \rightarrow b/c\_\_b$

$\Sigma^* reverse(\rho)$ = 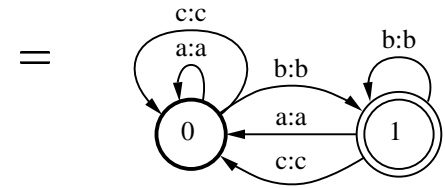

$Marker(\Sigma^* reverse(\rho), 1, \{>\}, \emptyset)$ = 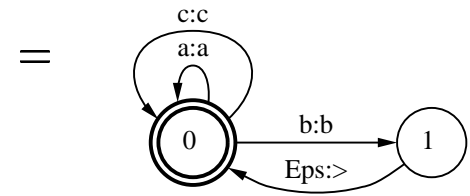

$reverse(Marker(\Sigma^* reverse(\rho), 1, \{>\}, \emptyset))$ = 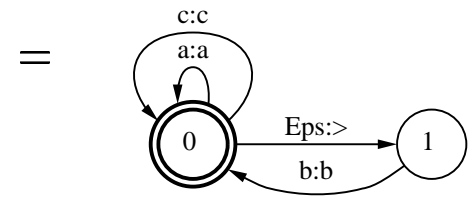

# The *Replace* Transducer

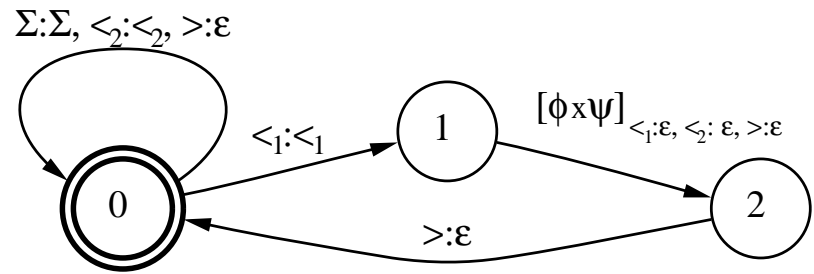

# Extension to Weighted Rules

Weighted context-dependent rules:

$$\phi \to \psi/\lambda \underline{\phantom{xx}} \rho$$

- $\phi, \lambda, \rho$ regular expressions,
- $\psi$ formal power series on the tropical semiring

Example:

$$c \to (.9c) + (.1t)/a \underline{\phantom{xx}} t$$

# Rational power series

Functions $S : \Sigma^* \to \mathcal{R}_+ \cup \{\infty\}$, *Rational* power series

- *Tropical semiring*: $(\mathcal{R}_+ \cup \{\infty\}, \min, +)$
- Notation: $S = \displaystyle\sum_{w \in \Sigma^*} (S, w)$
- Example: $S = (2a)(3b)(4b)(5b) + (5a)(3b)^*$
  $(S, abbb) = \min\{2+3+4+5 = 14, 5+3+3+3 = 11\} = 11$

**Theorem 6** *(Schützenberger, 1961): $S$ is rational iff it is recognizable (representable by a weighted transducer).*

# Compilation of weighted rules

- Extension of the composition algorithm to the weighted case
  - Efficient filter for $\epsilon$-transitions

  - Addition of weights of matching labels

- Same compilation algorithm

- Single-source shortest paths algorithms to find the best path

# Rewrite Rules: An Example

$$s \rightarrow z \,/\, \underline{\phantom{xx}} \,(\$|\#)\, \text{VStop} \,;$$

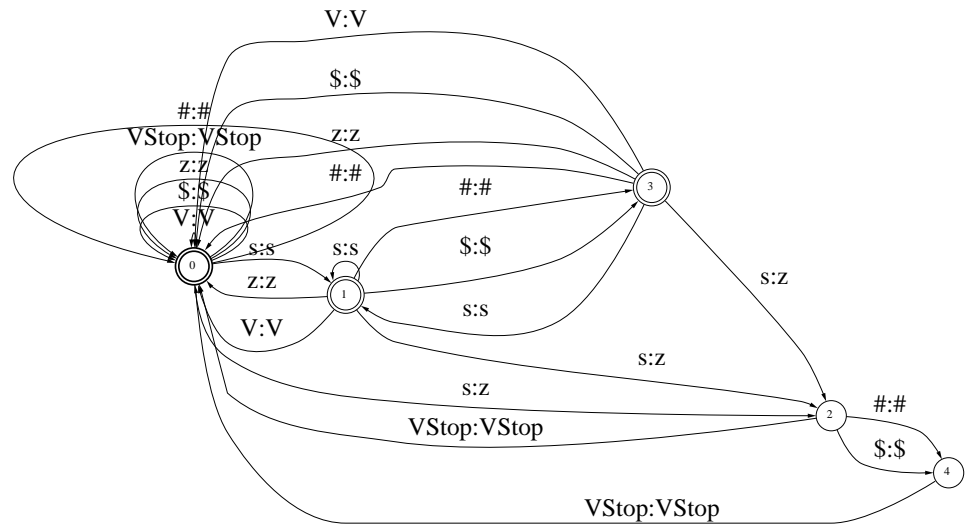

/mis\$mo\$/ ∘ Voicing = /miz\$mo\$/

# Syllable structure

[1] $(C^* \, V \, C^*_{1.0} \, \$ \,)^+ \cap$

[2] $\neg \, (\Sigma^* \, \$ \, (CC \cap \neg \, (CG \cup OL)) \, \Sigma^*) \cap$

[3] $\neg \, (\Sigma^* \, \$ \, ([+cor] \cup /x/ \cup /\beta/) \, /l/ \, \Sigma^*) \cap$

[4] $\neg \, (\Sigma^* \, \$ \, ([+cor,+strid] \cup /x/ \cup /\beta/) \, /r/ \, \Sigma^*)$

*estrella*: /estreya/ ∘ Intro( $ ) ∘ Syl ⇒

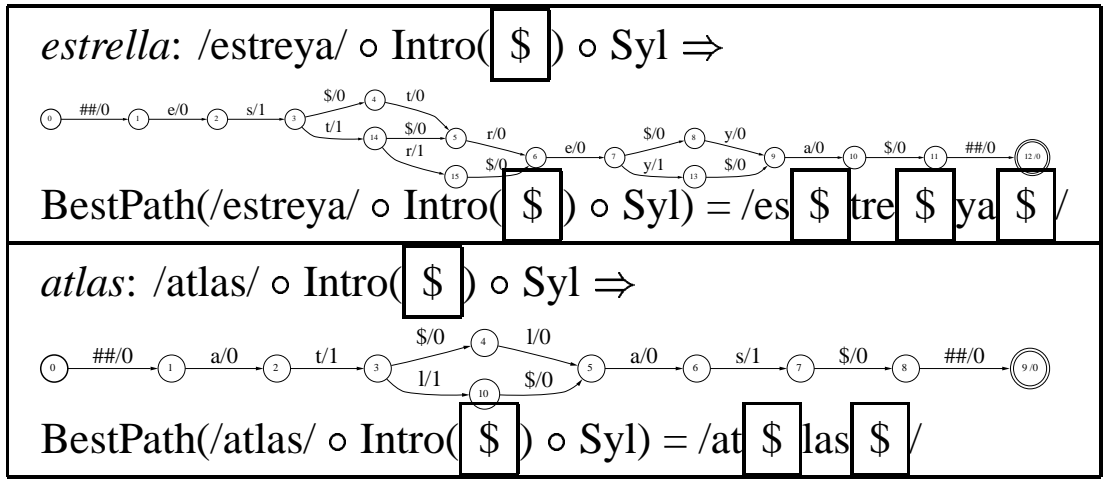

BestPath(/estreya/ ∘ Intro( $ ) ∘ Syl) = /es $ tre $ ya $ /

*atlas*: /atlas/ ∘ Intro( $ ) ∘ Syl ⇒

BestPath(/atlas/ ∘ Intro( $ ) ∘ Syl) = /at $ las $ /

# Russian Percentage Expansion: An example

с 5% скидкой

∘

| Lexical Analysis FST |

⇓

$s_{prep}$ $pjat_{num}$'$_{nom}$-$procentn_{adj}$ ⋆ +$aja_{fem+sg+nom}$ $skidk_{fem}oj_{sg+instr}$ ∪

$s_{prep}$ $pjat_{num}i_{gen}$-$procentn_{adj}$ ⋆ +$oj_{fem+sg+instr}$ $skidk_{fem}oj_{sg+instr}$ 2.0 ∪

$s_{prep}$ $pjat_{num}$'$ju_{instr}$-$procent_{noun}$+$ami_{pl+instr}$ $skidk_{fem}oj_{sg+instr}$ 4.0 ∪

⋮

○

**Language Model FSTs:**

$$\epsilon \rightarrow \boxed{\star} \ / \ \text{procent}_{\text{noun}} \ (\Sigma \cap \neg\#)^* \ \#\_\_ \ (\Sigma \cap \neg\#)^*_{\text{noun}}$$

$$\boxed{\star} \rightarrow \epsilon \ / \ \text{procentn}_{\text{adj}} \ (\Sigma \cap \neg\#)^* \ \#\_\_ \ (\Sigma \cap \neg\#)^*_{\text{noun}}$$

○

$$\epsilon \rightarrow \boxed{\star} \ / \ \text{procentn} \ (\Sigma \cap \neg\#)^*_{\text{Case} \cap \neg\text{instr}} \# \_\_ \ (\Sigma \cap \neg\#)^*_{\text{instr}}$$

$$\epsilon \rightarrow \boxed{\star} \ / \ \text{procentn} \ (\Sigma \cap \neg\#)^*_{\text{sg+Case}} \# \_\_ \ (\Sigma \cap \neg\#)^*_{\text{pl}}$$

○

$$\neg(\Sigma^* \boxed{\star} \Sigma^*)$$

⇓

$$\text{s pjati}_{\text{gen}}\text{-procentn}_{\text{adj}}\text{oj}_{\text{sg+instr}} \ \text{skidkoj}$$

## Percentage Expansion: Continued

с 5% скидкой

$\Downarrow$

s pjati$_{\text{gen}}$-procentn$_{\text{adj}}$oj$_{\text{sg+instr}}$ skidkoj

∘

$\boxed{\mathbf{L \circ P}}$

$\Downarrow$

```
s # PiT"!p~r@c"Entn&y # sK"!tk&y
```

# Phrasing Prediction

- **Problem**: predict intonational phrase boundaries in long unpunctuated utterences:

  *For his part, Clinton told reporters in Little Rock, Ark., on Wednesday || that the pact can be a good thing for America || if we change our economic policy || to rebuild American industry here at home || and if we get the kind of guarantees we need on environmental and labor standards in Mexico || and a real plan || to help the people who will be dislocated by it.*

- Bell Labs synthesizer uses a CART-based predictor trained on labeled corpora (Wang & Hirschberg 1992).

# Phrasing Prediction: Variables

For each $<w_i, w_j>$:

- length of utterance; distance of $w_i$ in syllables/ stressed syllables/words ... from the beginning/end of the sentence

- automatically predicted pitch accent for $w_i$ and $w_j$

- part-of-speech (POS) for a 4-word window around $<w_i, w_j>$;

- (largest syntactic constituent dominating $w_i$ but not $w_j$ and vice versa, and smallest constituent dominating them both)

- whether $<w_i, w_j>$ is dominated by an NP and, if so, distance of $w_i$ from the beginning of that NP, the NP, and distance/length

- (mutual information scores for a four-word window around $<w_i, w_j>$)

The most successful of these predictors so far appear to be POS, some constituency information, and mutual information

# Phrasing Prediction: Sample Tree

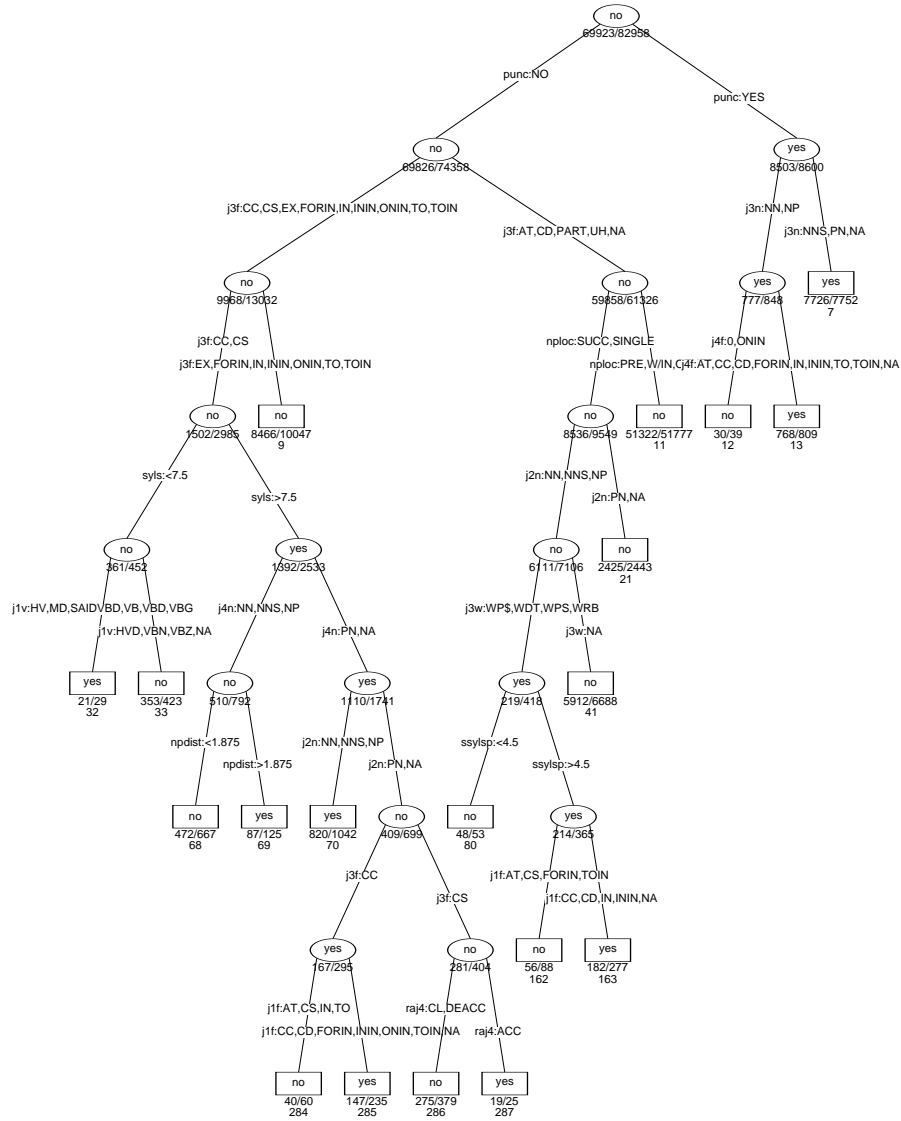

# Phrasing Prediction: Results

- Results for multi-speaker read speech:
  - major boundaries only: **91.2%**
  - collapsed major/minor phrases: **88.4%**
  - 3-way distinction between major, minor and null boundary: **81.9%**

- Results for spontaneous speech:
  - major boundaries only: **88.2%**
  - collapsed major/minor phrases: **84.4%**
  - 3-way distinction between major, minor and null boundary: **78.9%**

- Results for 85K words of hand-annotated text, cross-validated on training data: **95.4%**.

# Tree-Based Modeling: Prosodic Phrase Prediction

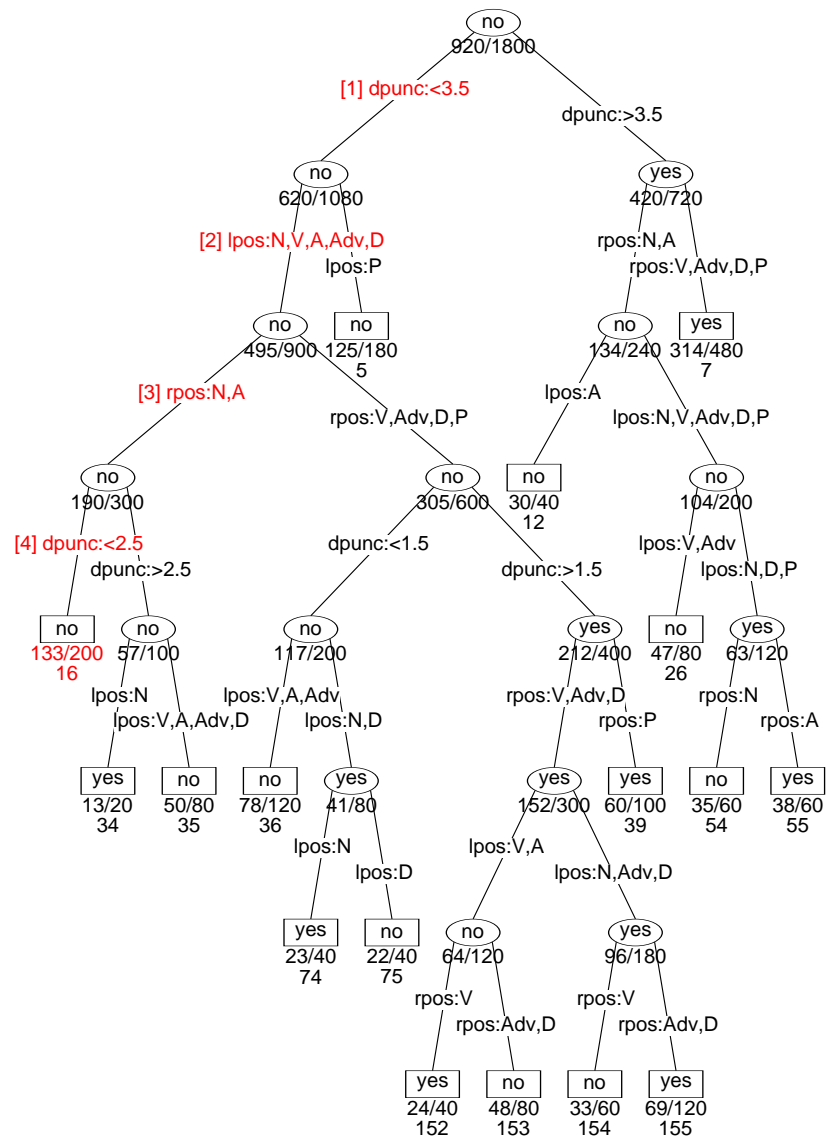

# The Tree Compilation Algorithm

(Sproat & Riley, 1996)

- Each *leaf* node corresponds to *single* rule defining a *constrained weighted mapping* for the input symbol associated with the tree

- Decisions at each node are stateable as regular expressions restricting the left or right context of the rule(s) dominated by the branch

- The full left/right context of the rule at a leaf node are derived by *intersecting* the expressions traversed between the root and leaf node

- The transducer for the entire tree represents the conjunction of all the constraints expressed at the leaf nodes; it is derived by *intersecting* together the set of WFSTs corresponding to each of the leaves
    - Note that intersection is defined for transducers that express same-length relations

- The *alphabet* is defined to be an alphabet of all correspondence pairs that were determined empirically to be possible

# Interpretation of Tree as a Ruleset

| Node 16 | $\lambda$ | | $\rho$ |
|---|---|---|---|
| | $\boxed{1}\ (\Sigma^*(I\omega \cup I\omega\#\omega \cup I\omega\#\omega\#\omega))\ \cap$ | | $\boxed{3}\ N \cup A$ |
| | $\boxed{2}\ (\Sigma^*(N \cup V \cup A \cup Adv \cup D))\ \cap$ | | |
| | $\boxed{4}\ (\Sigma^*(I\omega \cup I\omega\#\omega))$ | | |
| $\#\Rightarrow (I_{1.09} \cup \#_{0.41})\ /\ I(\omega\#)?(N \cup V \cup A \cup Adv \cup D)\ \underline{\quad}\ N \cup A$ ||||

# Summary of Compilation Algorithm

Each rule represents a *weighted two-level surface coercion rule*

$$Rule_L = Compile(\phi_T \to \psi_L / \bigcap_{p \in P_L} \lambda_p \underline{\quad} \bigcap_{p \in P_L} \rho_p)$$

Each tree/forest represents a set of simultaneous weighted two-level surface coercion rules

$$Rule_T = \bigcap_{L \in T} Rule_L$$

$$Rule_F = \bigcap_{T \in F} Rule_T$$

---

BestPath(,D#N#V#Adv#D#A#N ∘ Tree) ⇒ ,D#N#V#Adv$_{\Box,\Box}$D#A#N$_{2.76}$

# Lexical Ambiguity Resolution

- *Word sense disambiguation*:

  She handed down a harsh **sentence**.    *peine*

  This **sentence** is ungrammatical.    *phrase*

- *Homograph disambiguation*:

  He plays **bass**.                /be$^j$s/

  This lake contains a lot of **bass**.    /bæs/

- *Diacritic restoration*:

  appeler l'autre **cote** de l'atlantique    côté 'side'

  **Cote** d'Azur                côte 'coast'

(Yarowsky, 1992; Yarowsky 1996; Sproat, Hirschberg & Yarowsky, 1992; Hearst 1991)

# Homograph Disambiguation 1

- **N-Grams**

| Evidence | lɛd | lid | Logprob |
|---|---|---|---|
| lead *level/N* | 219 | 0 | 11.10 |
| *of* lead *in* | 162 | 0 | 10.66 |
| *the* lead *in* | 0 | 301 | 10.59 |
| lead *poisoning* | 110 | 0 | 10.16 |
| lead *role* | 0 | 285 | 10.51 |
| *narrow* lead | 0 | 70 | 8.49 |

- **Predicate-Argument Relationships**

| | | | |
|---|---|---|---|
| *follow/V* + lead | 0 | 527 | 11.40 |
| *take/V* + lead | 1 | 665 | 7.76 |

- **Wide Context**

| | | | |
|---|---|---|---|
| *zinc* ↔ lead | 235 | 0 | 11.20 |
| *copper* ↔ lead | 130 | 0 | 10.35 |

- **Other Features (e.g. Capitalization)**

# Homograph Disambiguation 2

Sort by $Abs(Log(\frac{Pr(Pron_1|Collocation_i)}{Pr(Pron_2|Collocation_i)}))$

| **Decision List for** *lead* | | |
|---|---|---|
| Logprob | Evidence | Pronunciation |
| 11.40 | *follow/V* + lead | ⇒ lid |
| 11.20 | *zinc* ↔ lead | ⇒ lɛd |
| 11.10 | lead *level/N* | ⇒ lɛd |
| 10.66 | *of* lead *in* | ⇒ lɛd |
| 10.59 | *the* lead *in* | ⇒ lid |
| 10.51 | lead *role* | ⇒ lid |
| 10.35 | *copper* ↔ lead | ⇒ lɛd |
| 10.28 | lead *time* | ⇒ lid |
| 10.16 | lead *poisoning* | ⇒ lɛd |

# Homograph Disambiguation 3: Pruning

- **Redundancy by subsumption**

| Evidence | lid | lɛd | Logprob |
|---|---|---|---|
| lead *level/N* | 219 | 0 | 11.10 |
| lead *levels* | 167 | 0 | 10.66 |
| lead *level* | 52 | 0 | 8.93 |

- **Redundancy by association**

| Evidence | tɛɚ | tɪɚ |
|---|---|---|
| tear *gas* | 0 | 1671 |
| tear ↔ *police* | 0 | 286 |
| tear ↔ *riot* | 0 | 78 |
| tear ↔ *protesters* | 0 | 71 |

# Homograph Disambiguation 4: Use

Choose single best piece of matching evidence.

| Decision List for *lead* | | |
|---|---|---|
| Logprob | Evidence | Pronunciation |
| 11.40 | *follow/V* + lead | ⇒ lid |
| 11.20 | *zinc* ↔ lead | ⇒ lɛd |
| 11.10 | lead *level/N* | ⇒ lɛd |
| 10.66 | *of* lead *in* | ⇒ lɛd |
| 10.59 | *the* lead *in* | ⇒ lid |
| 10.51 | lead *role* | ⇒ lid |
| 10.35 | *copper* ↔ lead | ⇒ lɛd |
| 10.28 | lead *time* | ⇒ lid |
| 10.16 | lead *poisoning* | ⇒ lɛd |

## Homograph Disambiguation: Evaluation

| Word | Pron1 | Pron2 | Sample Size | Prior | Performance |
|---|---|---|---|---|---|
| lives | laɪvz | lɪvz | 33186 | .69 | .98 |
| wound | waʊnd | wund | 4483 | .55 | .98 |
| Nice | naɪs | nis | 573 | .56 | .94 |
| Begin | bɪˈgɪn | beɪgɪn | 1143 | .75 | .97 |
| Chi | tʃi | kaɪ | 1288 | .53 | .98 |
| Colon | koʊˈloʊn | ˈkoʊlən | 1984 | .69 | .98 |
| lead (N) | lid | lɛd | 12165 | .66 | .98 |
| tear (N) | tɛɚ | tɪɚ | 2271 | .88 | .97 |
| axes (N) | ˈæksiz | ˈæksɪz | 1344 | .72 | .96 |
| IV | aɪ vi | fɔɹθ | 1442 | .76 | .98 |
| Jan | dʒæn | jɑn | 1327 | .90 | .98 |
| routed | ɹutɪd | ɹaʊtɪd | 589 | .60 | .94 |
| bass | beɪs | bæs | 1865 | .57 | .99 |
| TOTAL | | | 63660 | .67 | .97 |

# Decision Lists: Summary

- Efficient and flexible use of data.

- Easy to interpret and modify.

# Decision Lists as WFSTs

## The *lead* example

- Construct 'homograph taggers' $H_0, H_1 \ldots$ that find and tag instances of a homograph set in a lexical analysis. For example, $H_1$ is:

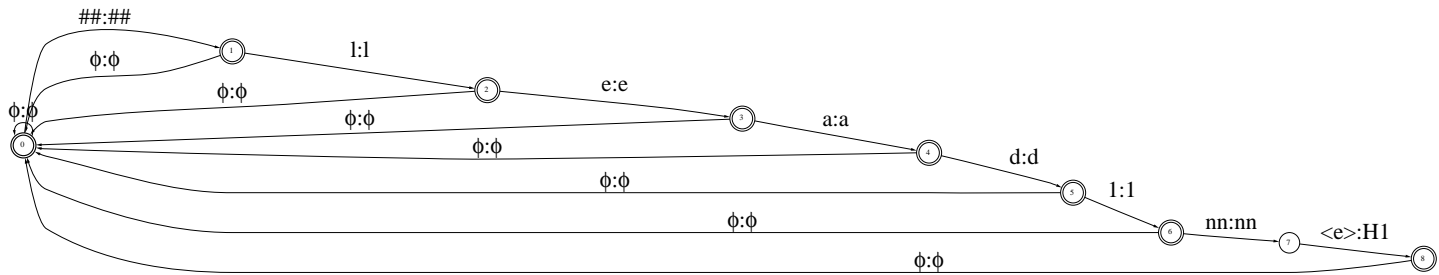

# Decision Lists as WFSTs

- Construct an *environmental classifier* consisting of a pair of transducers $C_1$ and $C_2$, where

    - $C_1$ optionally rewrites any symbol except the word boundary or the homograph tags H0, H1 ..., as a single dummy symbol $\Delta$

    - $C_2$ classifies contextual evidence from the decision list according to its type, and assigns a cost equal to the position of the evidence in the list; and otherwise passes $\Delta$, word boundary and H0, H1 ... through:

| | | |
|---|---|---|
| ## follow vb ## | $\rightarrow$ | ## $\Delta$ V0 ## $<1>$ |
| ## zinc nn ## | $\rightarrow$ | ## $\Delta$ C1 ## $<2>$ |
| ## level(s?) nn ## | $\rightarrow$ | ## $\Delta$ R1 ## $<3>$ |
| ## of pp ## | $\rightarrow$ | ## $\Delta$ [1 ## $<2>$ |
| ## in pp ## | $\rightarrow$ | ## $\Delta$ 1] ## $<2>$ |
| $\vdots$ | | |

# Decision Lists as WFSTs

- Construct a *disambiguator* $D$ from a set of *optional* rules of the form:

$$
\begin{aligned}
H0 &\rightarrow \diamond & / \quad & V0\ \Sigma^*\ \_\_ \\
H1 &\rightarrow \diamond & / \quad & C1\ \Sigma^*\ \_\_ \\
H1 &\rightarrow \diamond & / \quad & \_\_\ \Sigma^*\ C1 \\
H0 &\rightarrow \diamond & / \quad & \_\_\ \#\#\ \Delta^*\ R0 \\
H1 &\rightarrow \diamond & / \quad & \_\_\ \#\#\ \Delta^*\ R1 \\
H0 &\rightarrow \diamond & / \quad & [0\ \#\#\ \Delta^*\ \_\_\ \#\#\ \Delta^*\ 0] \\
H1 &\rightarrow \diamond & / \quad & [1\#\#\ \Delta^*\ \_\_\ \#\#\ \Delta^*\ 1] \\
&\vdots & & \\
H0 &\rightarrow \diamond <20> & & \\
H1 &\rightarrow \diamond <40> & &
\end{aligned}
$$

- Construct a *filter* $F$ that removes all paths containing H0, H1 . . . .

# Decision Lists as WFSTs

- Let an example input $T$ be:

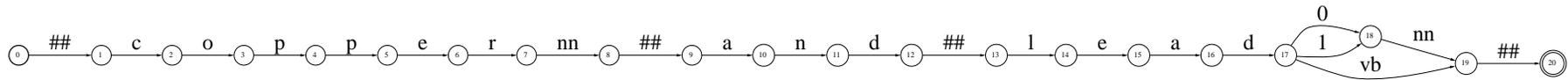

- Then the disambiguated input $T'$ is given by:

$$T \cap Project^{-1}[\, BestPath\, [\, T \circ H_0 \circ H_1 \circ C_1 \circ C_2 \circ D \circ F\, ]\, ]$$

# Syntactic Parsing and Analysis

- Intersection grammars (Voutilainen, 1994, inter alia)

- FST simulation of top-down parsing (Roche, 1996)

- Local grammars implemented as failure function automata (Mohri, 1994)

# Intersection Grammars

- **Text automaton** consisting of all possible lexical analyses of the input, including analysis of boundaries.

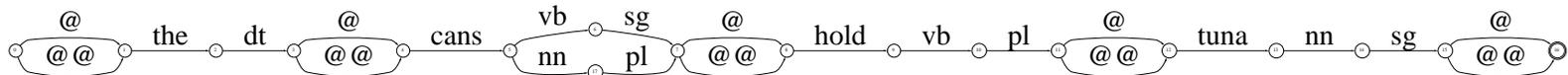

- Series of syntactic FSAs to be intersected with the text automaton, constraining it.

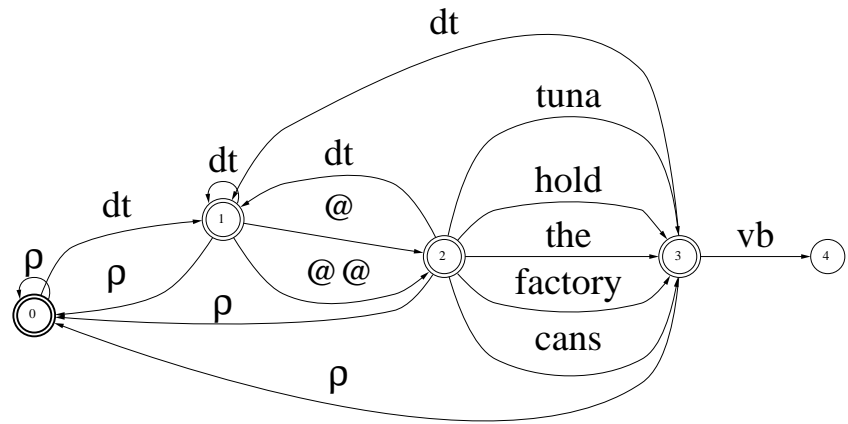

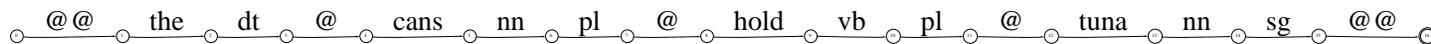

- Experimental grammars with a couple of hundred rules have been constructed.

# Top Down Parsing

$S \quad = \quad$ [S the cans hold tuna S]

$T_{dic} \quad =$

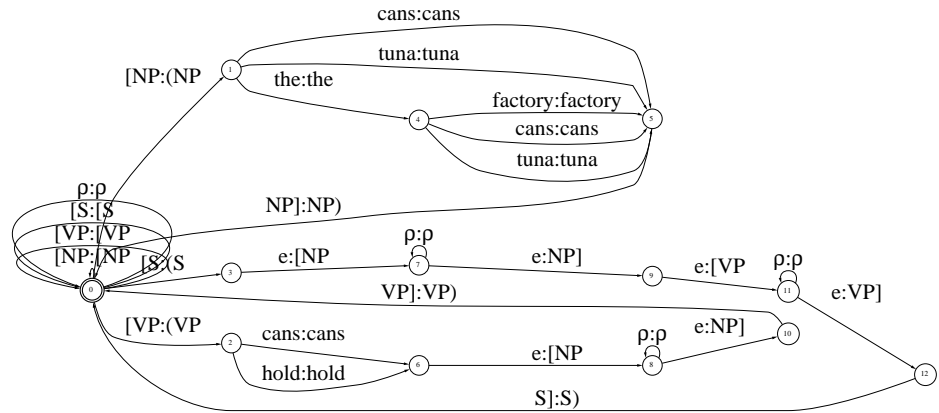

$S \circ T_{dic} \quad = \quad$ (S [NP NP] [VP the cans hold tuna VP] S)

(S [NP the NP] [VP cans hold tuna VP] S)

(S [NP the cans NP] [VP hold tuna VP] S)

(S [NP the cans hold NP] [VP tuna VP] S)

(S [NP the cans hold tuna NP] [VP VP] S)

$S \circ T_{dic} \circ T_{dic} \quad = \quad$ (S (NP the cans NP) (VP hold [NP tuna NP] VP) S)

# Local Grammars

- Descriptions of *local* syntactic phenomena, compiled into efficient, compact deterministic automata, using failure functions. (Cf. the use of failure functions with (sets of) *strings* familiar from string matching — e.g. Crochemore & Rytter, 1994)

- Descriptions may be *negative* or *positive*.
  Example of a negative constraint:
  - Let $L(G) = $ DT @ WORD VB
  - Construct deterministic automaton for $\Sigma^* L(G)$

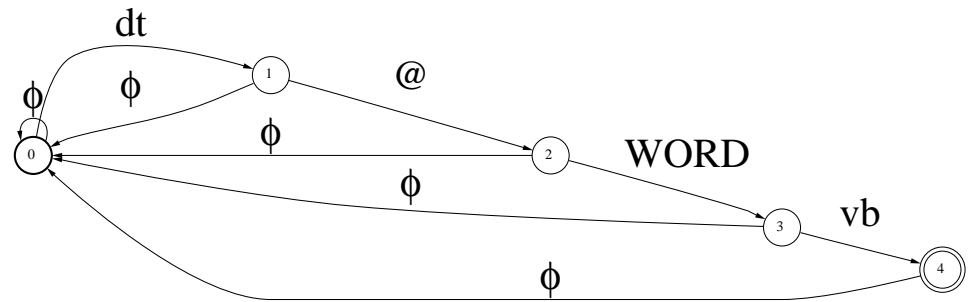

  - Given a sentence $L(S)$, compute $L(S) - \Sigma^* L(G) \Sigma^*$

# Indexation of natural language texts

- Motivation
  - Use of linguistic knowledge in indexation
  - Optimal complexities
    * Preprocessing of a text $t$, $O(|t|)$
    * Search for positions of a string $x$, $O(|x| + NumOccurrences(x))$

- Existing efficient indexation algorithms (*PAT*), but not convenient (use with large linguistic information)

# Example (1)

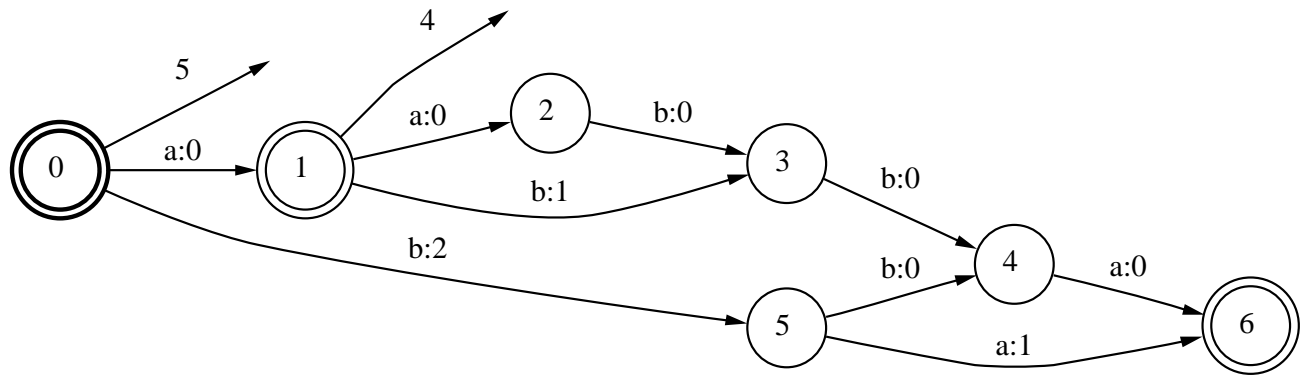

Figure 27: Indexation with subsequential transducers $t = aabba$.

# Example (2)

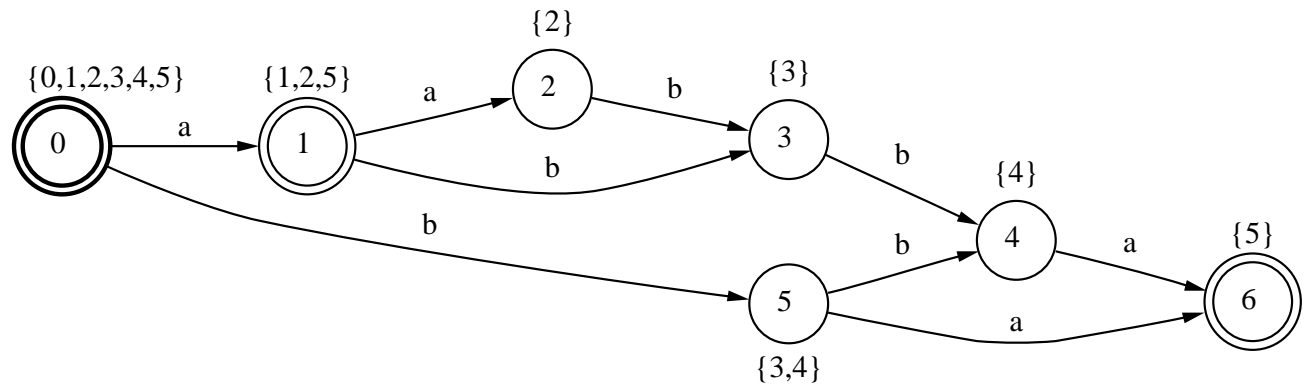

Figure 28: Indexation with automata $t = aabba$.

# Algorithms

- Based on the definition of an equivalence relation $R$ on $\Sigma^*$:
  $(w_1 \; R \; w_2)$ iff $w_1$ and $w_2$ have the same set of ending positions in $t$

- Construction
  - Minimal machines (subsequential transducer or automaton)
  - Use of a failure function to distinguish equivalence classes

- Can be adapted to natural language text
  (not storing list of positions of short words)

# Indexation with finite-state machines

- Complexity

  - Transducers (Crochemore, 1986): Preprocessing $O(|t|)$, Search $O(|x| + NumOccurrences(x))$ if using complex labels

  - Automata (Mohri, 1996b): Preprocessing quadratic, Search $O(|x| + NumOccurrences(x))$

- Advantage: use of linguistic information

  - Extended search: composition with morphological transducer

  - Refinement: composition with finite-state grammar

- Applications to WWW (Internet)

disambiguation. In *Proceedings of the thirteenth International Conference on Computational Linguistics (COLING'90), Helsinki, Finland*. COLING.

[34] Kuich, Wener and Arto Salomaa. 1986. *Semirings, Automata, Languages*. Springer-Verlag: Berlin-New York.

[35] Lee, C. and L. Rabiner. 1989. A frame-synchronous network search algorithm for connected word recognition *IEEE Transactions on ASSP* 37(11) 1649-1658.

[36] Lee, K.F. 1990. Context dependent phonetic hidden Markov models for continuous speech recognition. *IEEE Transactions on ASSP*. 38(4) 599–609.

[37] van Leeuwen, Hugo. 1989. *Too$_L$iP: A Development Tool for Linguistic Rules*. PhD thesis. Technical University Eindhoven.

[38] Leggestter, C.J. and P.C. Woodland. 1994. Speaker adaptation of continuous density HMMs using linear regression. *Proc. ICSLP '94* 2:

[73] Schützenberger, Marcel Paul. 1961. A remark on finite transducers. *Information and Control*, 4:185–196.

[74] Schützenberger, Marcel Paul. 1961. On the definition of a family of automata. *Information and Control*, 4.

[75] Schützenberger, Marcel Paul. 1977. Sur une variante des fonctions séquentielles. *Theoretical Computer Science*.

[76] Schützenberger, Marcel Paul. 1987. Polynomial decomposition of rational functions. In *Lecture Notes in Computer Science*, volume 386. Lecture Notes in Computer Science, Springer-Verlag: Berlin Heidelberg New York.

[77] Silberztein, Max. 1993. *Dictionnaires électroniques et analyse automatique de textes: le système INTEX*. Masson: Paris, France.

[78] Richard, Sproat. 1995. A finite-state architecture for tokenization and grapheme-to-phoneme conversion in multilingual text analysis. In *Proceedings of the ACL SIGDAT Workshop, Dublin, Ireland*. ACL.
M.Mohri-M.Riley-R.Sproat    Algorithms for Speech Recognition and Language Processing    REFERENCES    186